\documentclass[Afour,sageh,times]{sagej}
\usepackage{moreverb,url}

\usepackage{array}
\usepackage{multirow}

\usepackage{algorithm}
\usepackage{algpseudocode}

\usepackage{graphicx}
\usepackage{xcolor}

\usepackage[colorlinks,bookmarksopen,bookmarksnumbered,citecolor=red,urlcolor=red]{hyperref}

\newcommand\BibTeX{{\rmfamily B\kern-.05em \textsc{i\kern-.025em b}\kern-.08em
T\kern-.1667em\lower.7ex\hbox{E}\kern-.125emX}}

\def\volumeyear{2024}

\usepackage{xspace}
\makeatletter
\DeclareRobustCommand\onedot{\futurelet\@let@token\@onedot}
\def\@onedot{\ifx\@let@token.\else.\null\fi\xspace}

\def\eg{\emph{e.g}\onedot} 
\def\ie{\emph{i.e}\onedot}

\makeatother

\usepackage{amsmath}

\newcommand{\Exp}{\mathrm{Exp}}

\newcommand{\mbf}[1]{\mathbf{#1}}

\begin{document}

\runninghead{Huai \textit{et~al.}}

\title{SNAIL Radar: A large-scale diverse benchmark for evaluating 4D-radar-based SLAM}

\author{Jianzhu Huai\affilnum{1}, Binliang Wang\affilnum{1,$\dagger$}, Yuan Zhuang\affilnum{1,2,$\dagger$}, 
Yiwen Chen\affilnum{1}, Qipeng Li\affilnum{1}, Yulong Han\affilnum{1}}

\affiliation{\affilnum{1}The state key Lab of Information Engineering in Surveying, Mapping And Remote Sensing (LIESMARS), Wuhan University, Hubei, China 430079\\
\affilnum{2}Wuhan Quantum Institute, Hubei, China 430074}

\corrauth{$\dagger$Binliang Wang and Yuan Zhuang, 710 Xinghu Bldg, Info Faculty,
Wuhan University,
Wuhan, Hubei,
430079, China.}
\email{blwang@whu.edu.cn, yuan.zhuang@whu.edu.cn}

\begin{abstract}
4D radars are increasingly favored for odometry and mapping of autonomous systems due to their robustness in harsh weather and dynamic environments. 
Existing datasets, however, often cover limited areas and are typically captured using a single platform. 
To address this gap, we present a diverse large-scale dataset specifically designed for 4D radar-based localization and mapping. 
This dataset was gathered using three different platforms: a handheld device, an e-bike, and an SUV, under a variety of environmental conditions, including clear days, nighttime, and heavy rain.
The data collection occurred from September 2023 to February 2024, encompassing diverse settings such as roads in a vegetated campus and tunnels on highways.
Each route was traversed multiple times to facilitate place recognition evaluations. 
The sensor suite included a 3D lidar, 4D radars, stereo cameras, consumer-grade IMUs, and a GNSS/INS system. 
Sensor data packets were synchronized to GNSS time using a two-step process including a convex-hull-based smoothing and a correlation-based correction.
The reference motion for the platforms was generated by registering lidar scans to a terrestrial laser scanner (TLS) point cloud map
by a lidar inertial sequential localizer which supports forward and backward processing.
The backward pass enables detailed quantitative and qualitative assessments of reference motion accuracy.
To demonstrate the dataset’s utility, we evaluated several state-of-the-art radar-based odometry and place recognition methods,
indicating existing challenges in radar-based SLAM.

\end{abstract}


\maketitle

\section{Introduction}
Traditionally, automotive radars are integral to vehicles for obstacle detection and avoidance \citep{caesarNuScenesMultimodalDataset2020}. 
In recent years, millimeter-wave radars have also been utilized for odometry and mapping \citep{harlowNewWaveRobotics2023} thanks to their robustness to adverse conditions.
Two types of radars are commonly used: scanning radars, which rotate to capture 360$^\circ$-view scans \citep{cenPreciseEgomotionEstimation2018},
and solid-state single-chip radars, typically with a horizontal field of view of approximately 120$^\circ$ \citep{caiAutoPlaceRobustPlace2022,zhuang4DIRIOM4D2023}. 
Single-chip radars offer higher capture frequencies and are less bulky compared to their scanning counterparts.
With the Doppler velocity, a single-chip radar can easily detect moving objects, thus addressing significant challenges encountered in odometry approaches based on cameras \citep{wangDymSLAM4DDynamic2021} or lidars \citep{liMultisensor2023}.
However, conventional automotive radars often have poor vertical resolution, leading to the blending of low and high objects within the same view.

Recent advances in single-chip 4D imaging radars have enabled accurate elevation measurement thanks to improved vertical resolution along with range, azimuth, and Doppler velocity (hence 4D).
Utilizing millimeter waves, 4D radars can easily distinguish moving objects and perceive normally in adverse conditions, including fog, rain, and snow.
These features make 4D radars appealing for autonomous systems like robots and cars, which must operate reliably in such dynamic and adverse environments.

Several datasets for single-chip radars have been released.
Among them, the extensive nuScenes dataset features data captured by five 3D (XY+Doppler) radars mounted on an automobile.
However, its sequences are limited to discrete 20-second clips with minimal overlap, making it less suitable for SLAM and place recognition tasks.
Despite the surge in research utilizing 4D radars for ego-motion estimation, place recognition, and area mapping,
existing public 4D radar datasets, \eg, \cite{kramerColoRadar2022,liOdomBeyondVisionIndoorMultimodal2022,choiMSCRAD4RROSbasedAutomotive2023,zhangNTU4DRadLM4DRadarCentric2023},
often fall short in aspects such as accuracy of reference trajectories, diversity in data collection platforms, geographic scope, and repetition of data acquisition on same routes.

To address these deficiencies, we introduce a curated dataset from a data collection initiative by the SNAIL group at Wuhan University, 
which began in August 2022. 
This dataset encompasses a wide array of data sequences collected over a year using diverse platforms including a handheld rig, an e-bike, and an SUV. 
Though sensor configurations slightly vary across platforms, 
each sensor rig generally includes a stereo camera, one or two 4D radars, a 3D lidar, one or two IMUs, and a GNSS/INS system.
The published sequences were collected in a range of environments from a densely vegetated university campus to highways with tunnels and viaducts, 
during clear or rainy days, and nighttime.

Considering the applications in simultaneous localization and mapping (SLAM), we provide sequences with reference trajectories generated using data from a terrestrial laser scanner (TLS).
Synchronization based on proven techniques and meticulous calibration ensure the reliability of our data. 
We believe this dataset will significantly aid in evaluating algorithms for odometry, mapping, and place recognition based on 4D radar point clouds. 
The dataset is made available under the Open Database License
\endnote{License: The SNAIL radar dataset is made available under the Open Database License: \url{http://opendatacommons.org/licenses/odbl/1.0/}.
Any rights in individual contents of the dataset are licensed under the Database Contents License: \url{http://opendatacommons.org/licenses/dbcl/1.0/}}, at the website \endnote{The dataset website \url{https://snail-radar.github.io/}}.

Our contributions are summarized as follows:
\begin{enumerate}
\item We release a large-scale diverse 4D radar dataset,
captured at multiple times over selected routes by three different platforms,
encompassing diverse environmental conditions including rainy days and nights, campus roads, and highways.
Tools for data loading, visualization, and conversion, along with calibration results, are also provided.
\item Rigorous procedures are proposed and employed to synchronize motion-related messages of all sensors and calibrate the extrinsic parameters between them.
The sync procedure starts with the lidar and GNSS sync in hardware, 
then all sensor times are mapped to GNSS times with lidar times as the bridge,
and finally, the constant time offsets between all motion-related message types are estimated.
The extrinsic parameters are initialized with manual measurements and refined through a correlation method.
\item A sequential localization pipeline is proposed to generate the reference poses for the provided sequences.
The stitched TLS point clouds are used as submaps in a lidar-inertial odometry (LIO) method which operates in forward or backward mode.
We propose a technique to reverse messages for backward lidar-based odometry.
The backward processing enables quantitative accuracy evaluation of the reference trajectory.
\item On the released dataset, we compare three recent radar-based odometry methods, and three recent radar-based place recognition methods, 
showing the challenges in using 4D radars for SLAM.
\end{enumerate}

The structure of the paper is as follows. Section \ref{sec:related} reviews existing radar datasets.
The procedure for obtaining 4D radar point clouds is briefly described in Section \ref{sec:background}.
Details of the dataset, including data collection procedures, sensor setup, and file formats, are provided in Section \ref{sec:dataset}. 
Reference trajectory generation, data synchronization, sensor calibration, and known issues are discussed in Sections \ref{sec:truth}, 
\ref{sec:sync}, \ref{sec:calib}, and \ref{sec:issues}, respectively.
The comparative studies on odometry and place recognition methods are given in Section \ref{sec:comparison}.
Finally, Section \ref{sec:conclusions} summarizes the paper.

\section{Related work}
\label{sec:related}

This section reviews existing radar datasets for odometry and mapping with an emphasis on 4D radars.
A comprehensive survey of radar datasets targeting various applications including object detection and tracking have been given in \cite{zhouDeepRadarPerception2022}.

For the three types of radars, spinning radars, classic 3D automotive radars, and the recent 4D automotive radars, a variety of datasets have been published.
A detailed comparison of our dataset to the existing datasets is given in Table~\ref{tab:datasets}.

With spinning radars, several research groups have released radar localization datasets, \eg, the urban Oxford radar robotcar dataset \citep{barnesOxfordRadarRobotcar2020},
the Oxford offroad radar dataset \citep{gaddOORDOxfordOffroad2024},
the labeled
RADIATE perception dataset \citep{sheenyRADIATERadarDataset2021}, 
the urban MulRan place recognition dataset \citep{kimMulRanMultimodalRange2020}, 
and the multi-season Boreas dataset \citep{burnettBoreasMultiSeason2023}.

For the traditional 3D radars, the EU long-term odometry dataset \citep{yanEU2020} employed a Continental ARS308 radar for tracking objects.
Also, the nuScenes dataset \citep{caesarNuScenesMultimodalDataset2020} provides sequences of 20 seconds long captured by five automotive radars.
Its accurate poses are obtained by a Monte Carlo localization method relative to the HD map using both lidar and odometry data.

For the recent 4D radars, several datasets have been released.
The OdomBeyondVision dataset \citep{liOdomBeyondVisionIndoorMultimodal2022} was collected by an MAV, a UGV, and a handheld platform in buildings. All three platforms include a FLIR thermal camera and up to three single-chip TI AWR1843 4D radars.
The NTU4RadLM dataset \citep{zhangNTU4DRadLM4DRadarCentric2023} used an Oculii Eagle 4D radar and
an iRay thermal camera mounted on a handcart and a car to capture data in clear weather conditions within a campus.
The RRxIO dataset \citep{doerRadarVisualInertial2021} included a few indoor and outdoor sequences acquired by a thermal camera and a IWR6843 radar mounted on an MAV.
The reference trajectories were generated by a motion capture system and a visual inertial SLAM system.
The Coloradar dataset \citep{kramerColoRadar2022} was captured by a handheld platform outfitted with a single-chip radar and a cascaded radar.
The dataset spans buildings, mine tunnels, and outdoor areas, with reference trajectories generated by a lidar SLAM system.
One of its imperfections is that some sequences have wrong Doppler measurements.
The USVInland dataset \citep{chengAreWeReady2021} was captured by an unmanned surface vehicles with three TI radars, traveling on inland waterways.
The sequences with good GNSS RTK solutions are provided for SLAM evaluations.
The MSC RAD4R \citep{choiMSCRAD4RROSbasedAutomotive2023} dataset consists of many urban and outskirt sequences captured by a sensor rig mounted on a car.
Though the sensor rig includes a GNSS/RTK system and a AHRS system, the RTK solutions often have big closure errors in height and the AHRS system gave wrong headings.

\begin{table*}[]
	\centering
	\caption{Our datasets in comparison to existing radar-based SLAM datasets. Our dataset involves three platforms and diverse
		scenes, covering small to large geographic scopes. RA: range azimuth, PC + D: point cloud and Doppler, ADC: analog to digital
		converter samples.}
	\label{tab:datasets}
	\begin{tabular}{ccccccc}
		\hline
		\textbf{Dataset}                                                     & \textbf{Radar}                                                      & \textbf{Data type}                                       & \textbf{Ground truth}                                            & \textbf{Platform}                                            & \textbf{Weather/Light}                                           & \textbf{Scenarios}                                                            \\ \hline
		\begin{tabular}[c]{@{}c@{}}Oxford Radar\\ RobotCar 2020\end{tabular} & Navtech                                                             & RA                                                       & \begin{tabular}[c]{@{}c@{}}PGO with\\ GPS/INS\end{tabular}       & SUV                                                          & \begin{tabular}[c]{@{}c@{}}fog, night,\\ rain, snow\end{tabular} & urban road                                                                    \\ \hline
		\begin{tabular}[c]{@{}c@{}}RADIATE\\ 2021\end{tabular}               & Navtech                                                             & RA                                                       & GPS/IMU                                                          & car                                                          & \begin{tabular}[c]{@{}c@{}}fog, night,\\ rain, snow\end{tabular} & urban road, park                                                              \\ \hline
		MulRan 2020                                                          & Navtech                                                             & RA                                                       & \begin{tabular}[c]{@{}c@{}}PGO with\\ GPS\end{tabular}           & car                                                          & clear                                                            & \begin{tabular}[c]{@{}c@{}}urban road, \\ tunnel, campus\end{tabular}         \\ \hline
		Boreas 2023                                                          & Navtech                                                             & RA                                                       & \begin{tabular}[c]{@{}c@{}}GNSS/IMU/\\ wheel fusion\end{tabular} & SUV                                                          & \begin{tabular}[c]{@{}c@{}}night, rain,\\ snow\end{tabular}      & urban road                                                                    \\ \hline
		OORD 2024                                                            & Navtech                                                             & RA                                                       & GPS                                                              & SUV                                                          & night, snow                                                      & off-road                                                                      \\ \hline
		\begin{tabular}[c]{@{}c@{}}EU Long-term\\ 2020\end{tabular}          & Conti ARS 308                                                       & 2D PC + D                                                & GPS-RTK/IMU                                                      & car                                                          & night, snow                                                      & urban road                                                                    \\ \hline
		\begin{tabular}[c]{@{}c@{}}ColoRadar\\ 2022\end{tabular}             & Two TI AWR                                                          & \begin{tabular}[c]{@{}c@{}}ADC,\\ 3D PC + D\end{tabular} & Vicon or PGO                                                     & handheld                                                     & clear                                                            & \begin{tabular}[c]{@{}c@{}}mine, office,\\ outdoors\end{tabular}              \\ \hline
		\begin{tabular}[c]{@{}c@{}}MSC-RAD4R\\ 2023\end{tabular}             & Oculii Eagle                                                        & 3D PC + D                                                & GPS-RTK                                                          & car                                                          & \begin{tabular}[c]{@{}c@{}}night, smoke,\\ snow\end{tabular}     & urban road                                                                    \\ \hline
		\begin{tabular}[c]{@{}c@{}}USVInland\\ 2021\end{tabular}             & TI AWR1843                                                          & 3D PC + D                                                & GPS-RTK/IMU                                                      & boat                                                         & clear                                                            & waterway                                                                      \\ \hline
		\begin{tabular}[c]{@{}c@{}}Odom Beyond\\ Vision 2022\end{tabular}    & TI AWR1843                                                          & 3D PC + D                                                & \begin{tabular}[c]{@{}c@{}}MoCap or\\ LOAM\end{tabular}          & \begin{tabular}[c]{@{}c@{}}handheld,\\ UAV, UGV\end{tabular} & smoke                                                            & indoors                                                                       \\ \hline
		\begin{tabular}[c]{@{}c@{}}NTU4DRadLM\\ 2023\end{tabular}            & Oculli Eagle                                                        & 3D PC + D                                                & \begin{tabular}[c]{@{}c@{}}LVI-SLAM +\\ PGO\end{tabular}         & car                                                          & clear                                                            & campus                                                                        \\ \hline
		Ours                                                                 & \begin{tabular}[c]{@{}c@{}}Oculli Eagle\\ Conti ARS548\end{tabular} & 3D PC + D                                                & \begin{tabular}[c]{@{}c@{}}LIO loc. on\\ TLS map\end{tabular}    & \begin{tabular}[c]{@{}c@{}}handheld,\\ SUV, UGV\end{tabular} & night, rain                                                      & \begin{tabular}[c]{@{}c@{}}campus, highway, \\ tunnels, overpass\end{tabular} \\ \hline
	\end{tabular}
\end{table*}

In view of existing datasets, our dataset covers diverse environments (\eg, vegetated campus, tunnels, rain, and nights) using three platforms, featuring repetitive traversals of selected paths and accurate reference solutions.

\section{Radar signal processing background}
\label{sec:background}
This section reviews the typical data acquisition and processing pipeline for the multi-input-multi-output (MIMO) frequency modulated continuous wave (FMCW) 4D radars,
mainly referring to the Texas Instruments (TI) products.

\begin{figure}[h!]
	\centering
	\includegraphics[width=0.95\columnwidth]{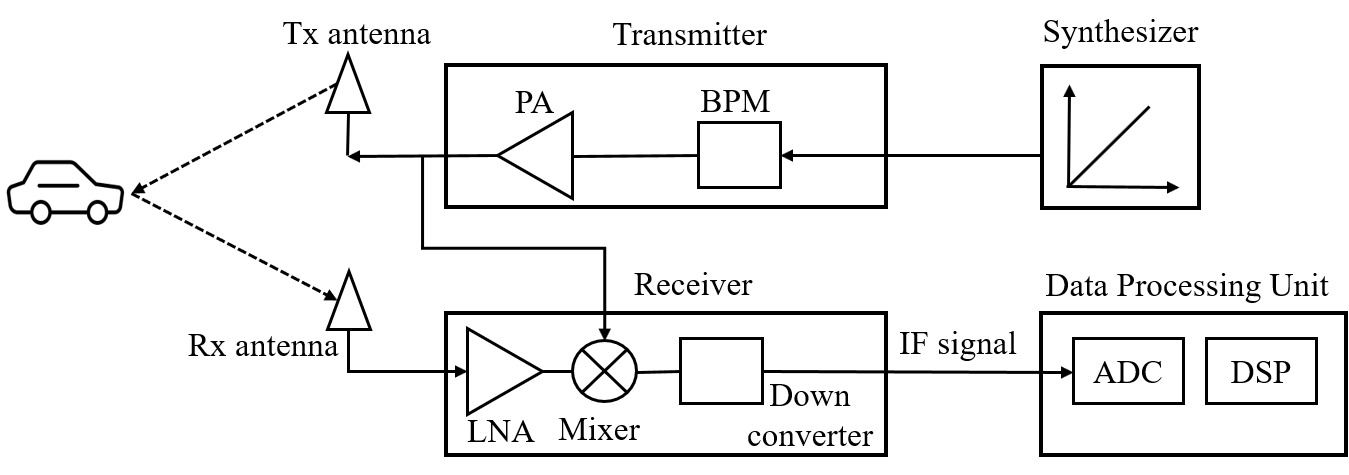}\\ \vspace{6pt}
	\includegraphics[width=0.95\columnwidth]{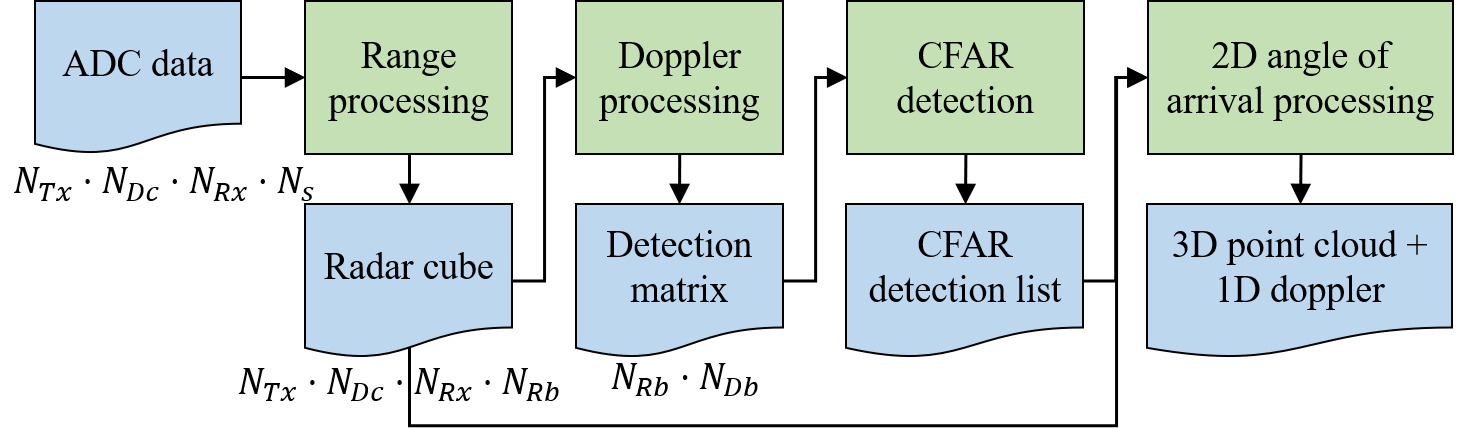}
	\caption{(Top) The schematic of the millimeter wave radar's mechanism, as in a typical 4D radar by Texas Instruments (TI) \citep{iovescuFundamentalsMillimeterWave2017}.
	PA: power amplifier, BPM: binary phase modulation, LNA: low noise amplifier, IF: intermediate frequency, ADC: analog-to-digital converter, DSP: digital signal processor, AMP: amplifier.
	(Bottom) The simplified radar point cloud generation pipeline from the ADC samples.
	CFAR: constant false alarm rate. The math symbols are explained in the main text.}
\label{fig:dsp}
\end{figure}

As shown in the top diagram of Fig.~\ref{fig:dsp}, the transmitter antennas send out radio frequency chirps with increasing frequency. 
The receiver antennas detect the waves reflected by objects and
downconvert the signals by mixing them with the transmitted carrier wave to obtain signals at the intermediate frequency.
These signals are then sampled by an analog-to-digital converter (ADC) to get the complex (in-phase and quadrature components, I/Q) ADC samples.
These ADC samples are passed to the digital signal processors (DSPs) for further processing.

The digital signal processing, depicted in the bottom of Fig.~\ref{fig:dsp}, includes four fundamental components: range processing, Doppler processing, 
constant false alarm rate (CFAR) detection, and 2D angle of arrival (AoA) processing.
Each component is thoroughly illustrated in the TI mmwave SDK documentation.
The mathematical principles underlying this pipeline are explained in \cite{iovescuFundamentalsMillimeterWave2017}.
Here, we provide a high-level functional description of these components, with advanced options omitted for clarity.

Let's denote the following:
\begin{itemize}
	\item $N_{Dc}$: number of Doppler chirps per frame
	\item $N_{Rx}$: number of receiver antennas
	\item $N_{Tx}$: number of transmitter antennas
	\item $N_s$: number of ADC samples per chirp duration
	\item $N_{Db}$: number of Doppler bins in the fast Fourier transform (FFT)
	\item $N_R$: number of range bins in the FFT
\end{itemize}

The range processing routine takes as input ADC chirp samples of I/Q values, 
performs 1D range FFT and optional direct current (DC) range calibration during the active frame time, 
and outputs a radar cube.
The ADC data has a size of $N_{Tx} \cdot N_{Dc} \cdot N_{Rx} \cdot N_s$.
The radar cube has a size of $N_{Tx} \cdot N_{Dc} \cdot N_{Rx} \cdot N_{Rb}$.

The Doppler processing routine takes the preceding radar cube,
performs 2D Doppler FFT and energy summation during the inter-frame time, and
outputs a detection matrix of size $N_{Rb} \cdot N_{Db}$.

The CFAR routine starts with the detection matrix, 
performs CFAR detection and peak grouping to output a CFAR detection list of objects in the range-Doppler domain.

The 2D AoA processing routine takes the radar cube and the CFAR detection list,
performs 2D Doppler FFT on relevant entries of the detected objects, followed by 
a 2D angle FFT and peak selection by CFAR to determine the azimuth and elevation for the detected objects,
resulting in 3D point clouds with 1D Doppler velocity.

\section{Dataset}
\label{sec:dataset}
Our dataset, dubbed snail-radar, is collected using three platforms, a handheld device, an e-bike, and an SUV.
The three platforms share almost the same sensor rig as the data collection was carried out sequentially across these platforms.
The published dataset includes 44 sequences: 6 handheld sequences, 14 e-bike sequences, and 24 SUV sequences.
For each sequence, Table~\ref{tab:seqs} specifies the platform, weather and lighting conditions, and the approximate traveled distance and duration.

\begin{table}[]
	\centering
	\caption{The 44 sequences of our dataset are repeatedly collected at 8 routes with three platforms in diverse weather/lighting conditions. The route shorthands bc=basketball court, sl=starlake, ss=software school, if=info faculty, iaf=info and arts faculty, iaef=info, arts, and engineering faculty, st=starlake tower, 81r=August 1 road. The platforms H=handheld, E=ebike, S=SUV. The weather and lighting condition is clear and daytime unless specified. Different to other sequences of the info and arts faculty, the 20240113/2 sequence's direction is anticlockwise.}
	\label{tab:seqs}
	\begin{tabular}{@{}ccccc@{}}
		\toprule
		\multicolumn{2}{c}{\textbf{\begin{tabular}[c]{@{}c@{}}Route\\ \&plat.\end{tabular}}} &
		\textbf{Date/Run} &
		\textbf{\begin{tabular}[c]{@{}c@{}}Dist.(m)/\\ Dur.(sec)\end{tabular}} &
		\textbf{\begin{tabular}[c]{@{}c@{}}Weather/\\ lighting\end{tabular}} \\ \midrule
		\multirow{5}{*}{bc}   & H & 20230920/1      & 84/74     & light rain, night \\
		& H & 20230921/2      & 59/83     & mod. rain         \\
		& E & 20231007/4      & 78/51     &                   \\
		& E & 20231105/6      & 251/144   & mod. rain         \\
		& E & 20231105\_aft/2 & 167/104   & light rain        \\ \midrule
		\multirow{9}{*}{sl}   & H & 20230920/2      & 2182/1839 & light rain. night \\
		& H & 20230921/3      & 2171/1857 & mod. rain         \\
		& H & 20230921/5      & 2015/1731 & mod. rain         \\
		& E & 20231007/2      & 1997/657  &                   \\
		& E & 20231019/1      & 1919/460  & night             \\
		& E & 20231105/2      & 2045/524  & heavy rain        \\
		& E & 20231105/3      & 2069/537  & heavy rain        \\
		& E & 20231105\_aft/4 & 2019/698  & light rain        \\
		& E & 20231109/3      & 1983/546  &                   \\ \midrule
		\multirow{6}{*}{ss}   & H & 20230921/4      & 736/622   & light rain        \\
		& E & 20231019/2      & 781/533   & night             \\
		& E & 20231105/4      & 895/395   & heavy rain        \\
		& E & 20231105/5      & 967/400   & mod. rain         \\
		& E & 20231105\_aft/5 & 826/534   & light rain        \\
		& E & 20231109/4      & 795/533   &                   \\ \midrule
		\multirow{7}{*}{if}   & S & 20231208/4      & 2228/515  &                   \\
		& S & 20231213/4      & 2227/494  & light rain. night \\
		& S & 20231213/5      & 2225/475  & light rain, night \\
		& S & 20240115/3      & 2223/525  & dusk              \\
		& S & 20240116/5      & 2228/514  & dusk              \\
		& S & 20240116\_eve/5 & 2224/462  & night             \\
		& S & 20240123/3      & 2231/535  &                   \\ \midrule
		\multirow{8}{*}{iaf}  & S & 20231201/2      & 4620/1042 &                   \\
		& S & 20231201/3      & 4631/946  &                   \\
		& S & 20231208/5      & 4616/870  &                   \\
		& S & 20231213/2      & 4603/938  & light rain, night \\
		& S & 20231213/3      & 4613/875  & light rain, night \\
		& S & 20240113/2      & 4610/1005 & anticlockwise     \\
		& S & 20240113/3      & 4613/962  &                   \\
		& S & 20240116\_eve/4 & 4609/868  & night             \\ \midrule
		\multirow{3}{*}{iaef} & S & 20240113/5      & 7283/1509 &                   \\
		& S & 20240115/2      & 6641/1374 &                   \\
		& S & 20240116/4      & 6648/1515 &                   \\ \midrule
		\multirow{3}{*}{st}   & S & 20231208/1      & 275/147   &                   \\
		& S & 20231213/1      & 276/126   & light rain, night \\
		& S & 20240113/1      & 541/214   &                   \\ \midrule
		\multirow{3}{*}{81r}  & S & 20240116/2      & 8554/1433 & light rain        \\
		& S & 20240116\_eve/3 & 8521/1293 & night             \\
		& S & 20240123/2      & 8539/1743 &                   \\ \bottomrule
	\end{tabular}
\end{table}

More than half of these sequences were captured under special conditions such as rain, dusk, and night.
The dataset encompasses various scenes including campus environments, highways, and tunnels.
Sample data from camera, lidar, and 4D radar are provided in Table~\ref{tab:samples}.
For place recognition purposes, data for each route were captured at least three different times.
Additionally, each sequence was recorded with roughly the same start and end poses.

\newcommand{\addpic}[1]{\includegraphics[width=0.23\textwidth]{#1}}
\newcolumntype{C}{>{\centering\arraybackslash}m{0.23\textwidth}}
\begin{table*}\sffamily
\centering
\begin{tabular}{l@{\hspace{5pt}}C@{\hspace{3pt}}C@{\hspace{3pt}}C@{\hspace{3pt}}C@{}}
\toprule
& \textbf{ZED2i Left} & \textbf{Hesai XT32} & \textbf{Oculii Eagle} & \textbf{ARS548} \\
\midrule
\textbf{D} & \addpic{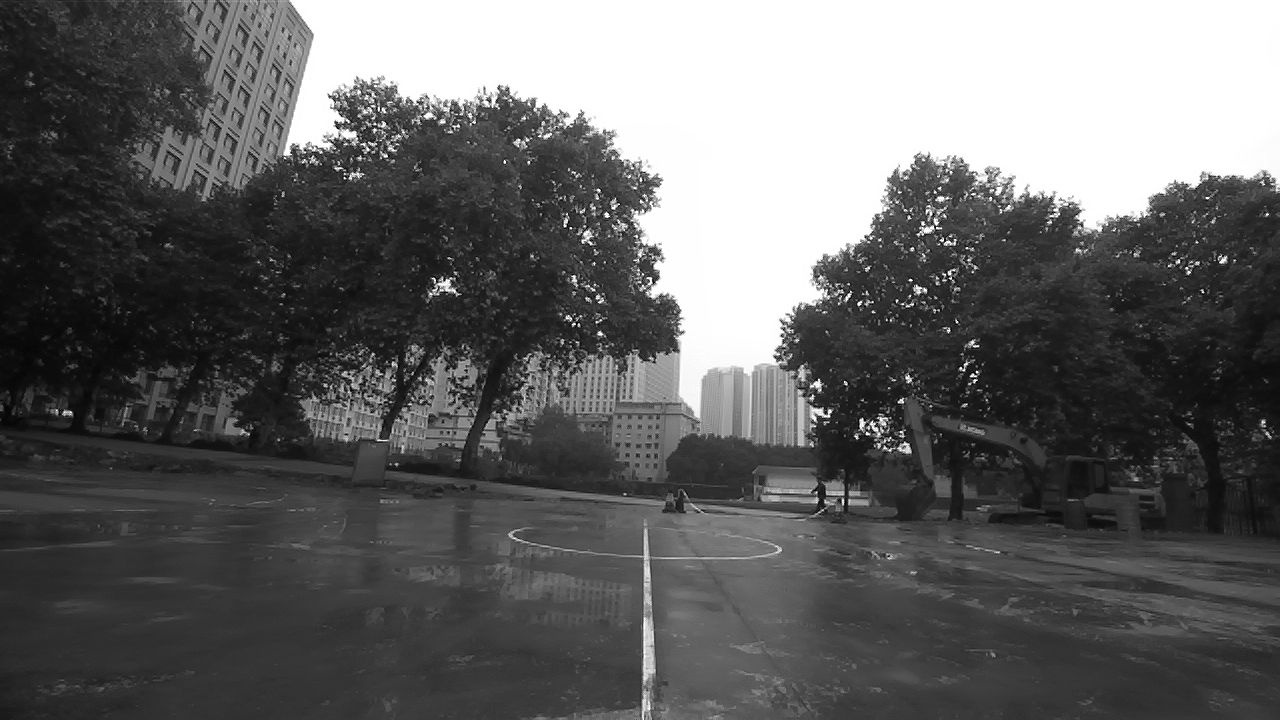} & \addpic{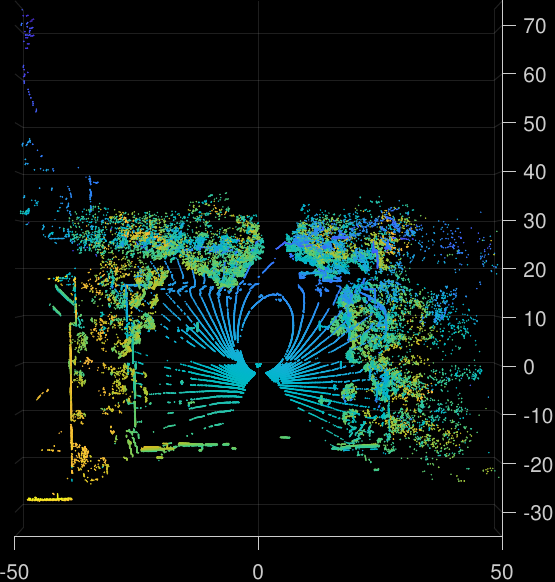} & \addpic{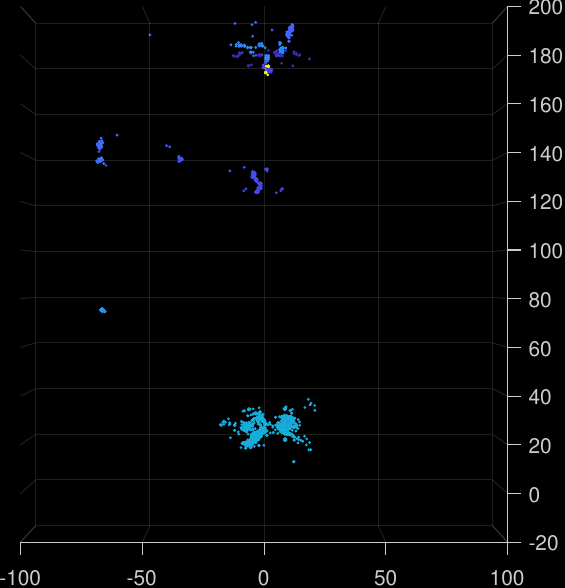} & \addpic{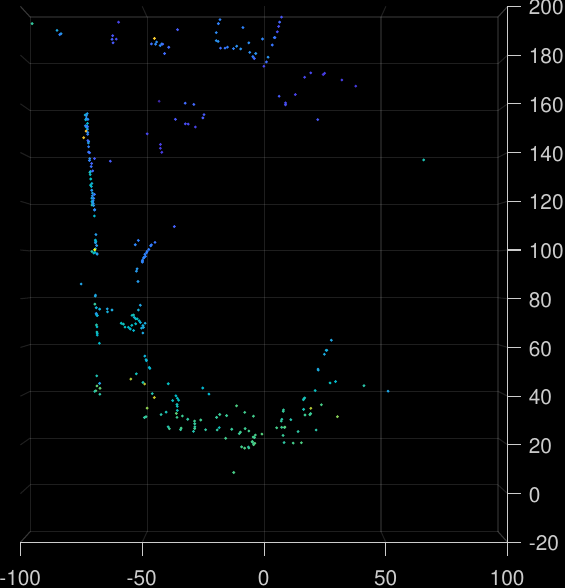} \\
\textbf{R} & \addpic{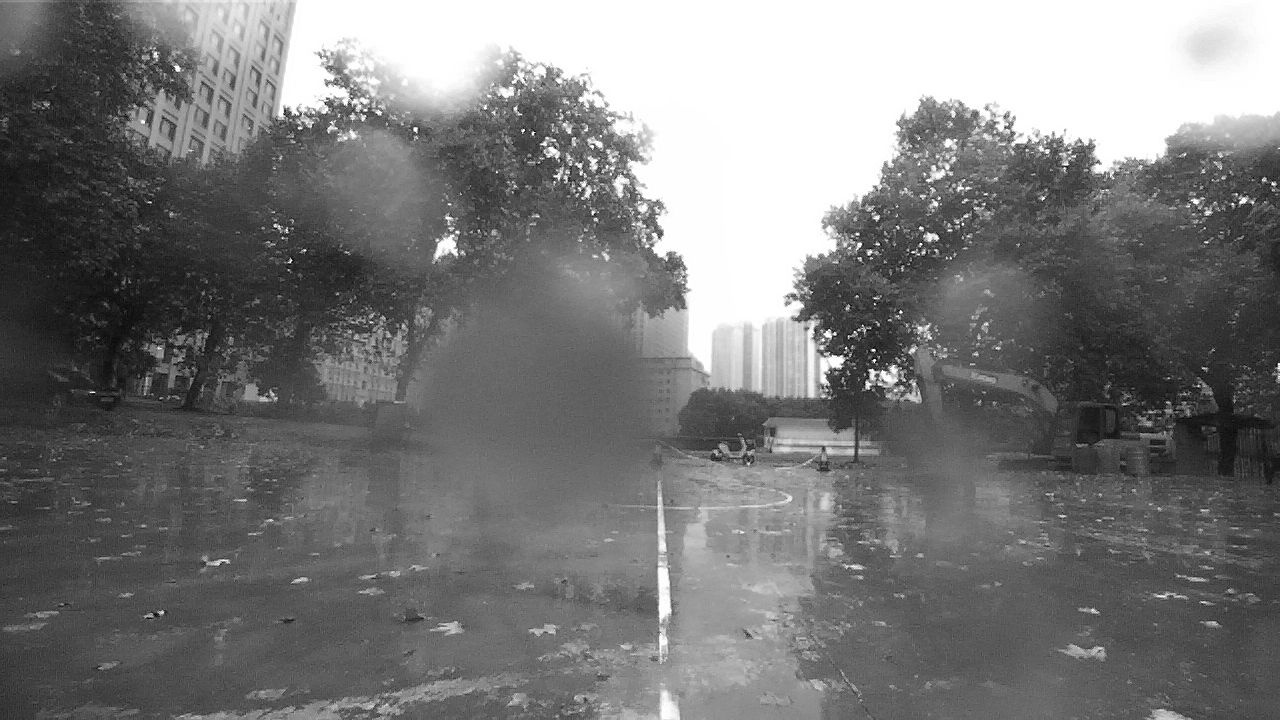} & \addpic{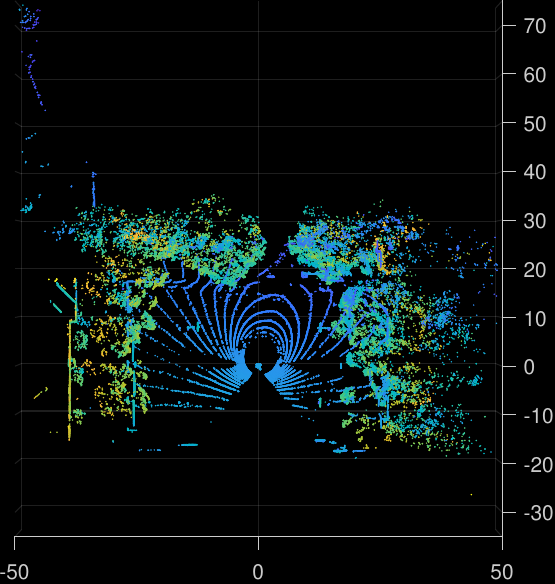} & \addpic{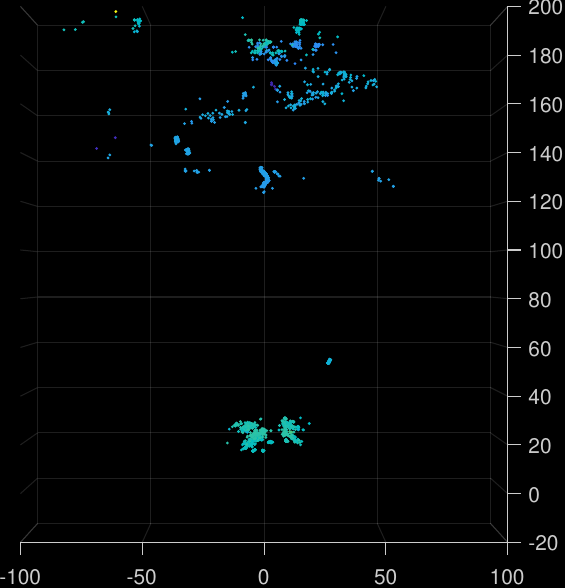} & \addpic{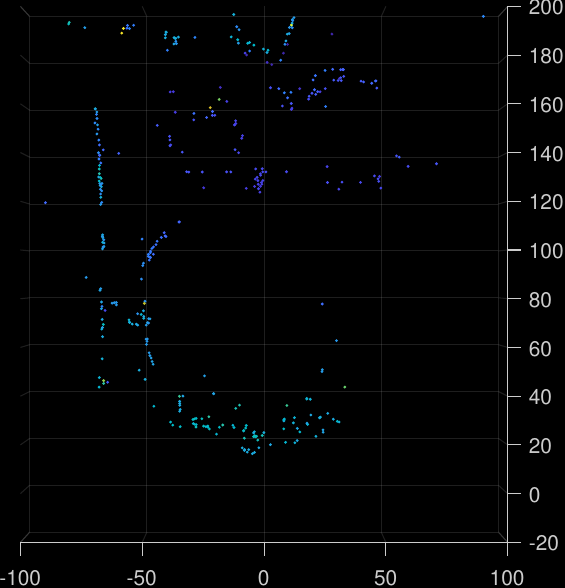} \\
\textbf{N} & \addpic{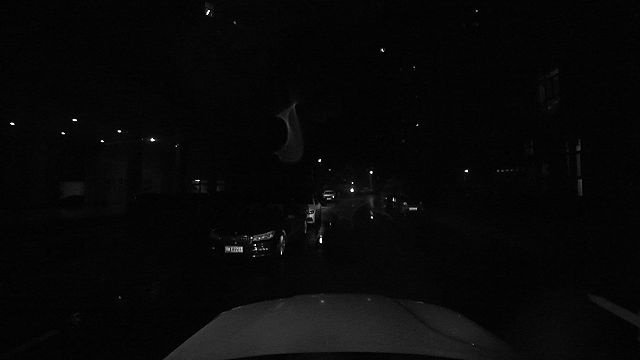} & \addpic{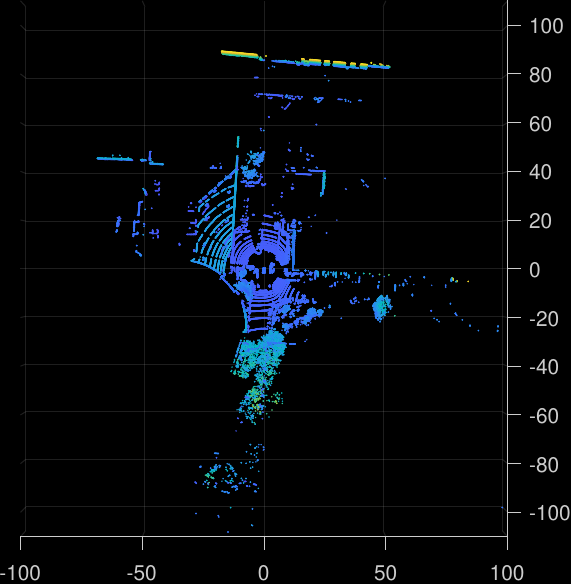} & \addpic{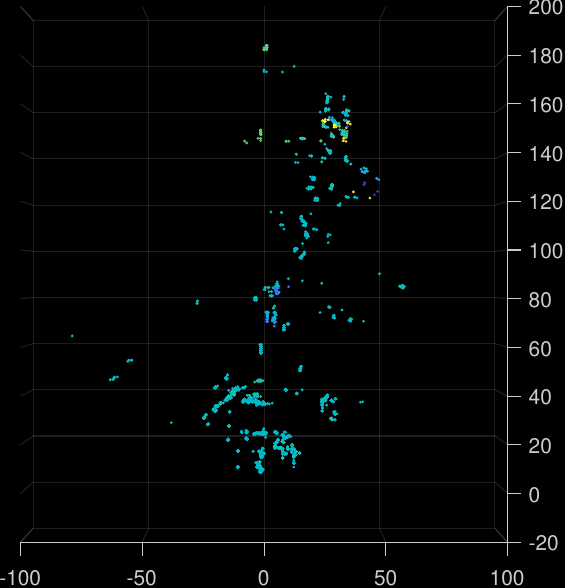} & \addpic{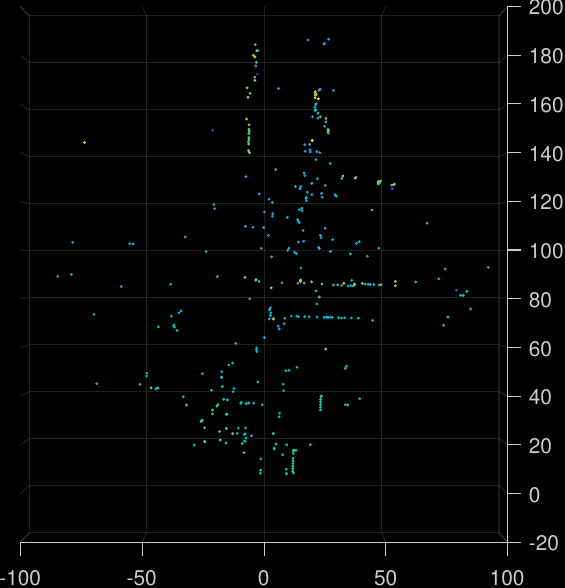} \\
\textbf{H} & \addpic{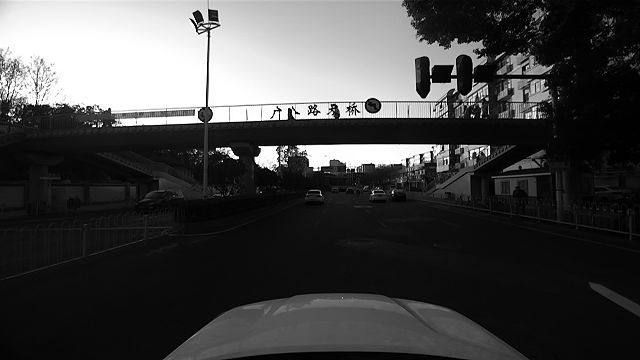} & \addpic{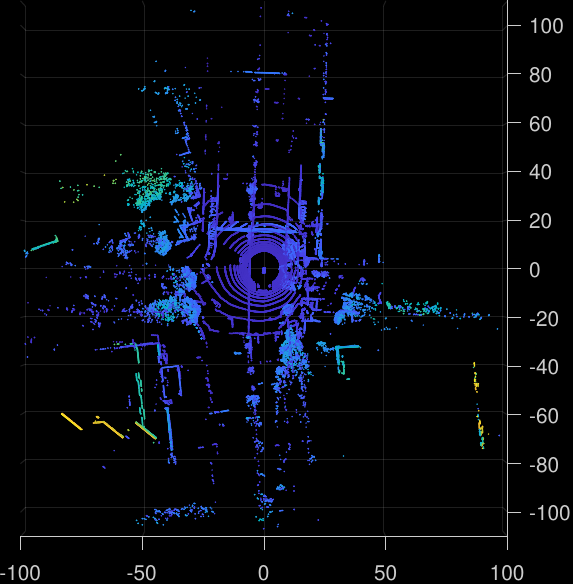} & \addpic{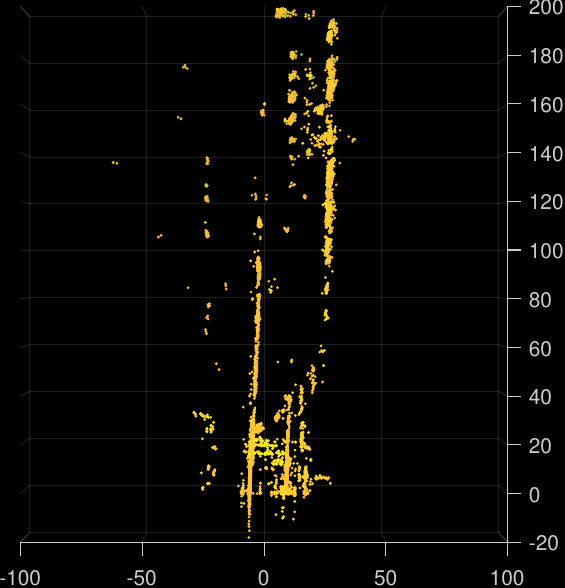} & \addpic{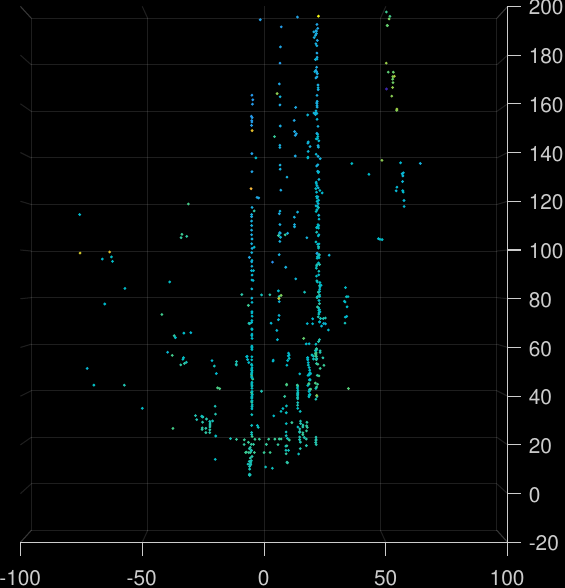} \\
\textbf{T} & \addpic{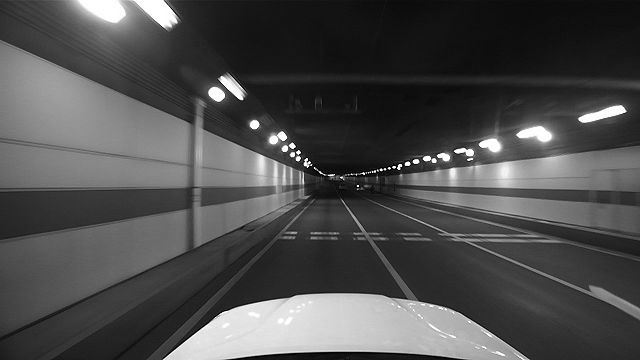} & \addpic{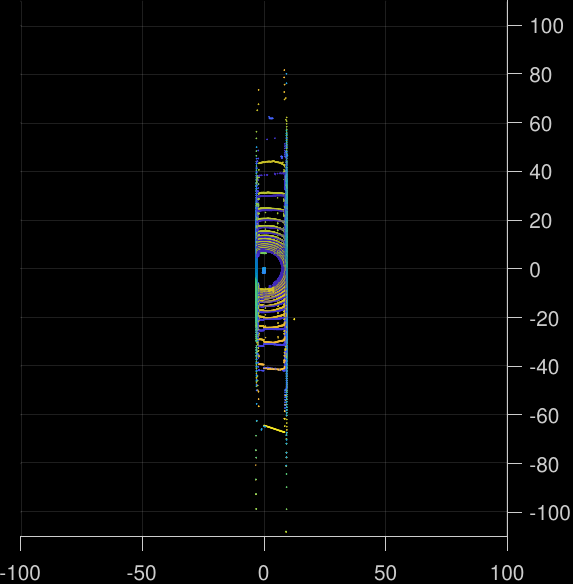} & \addpic{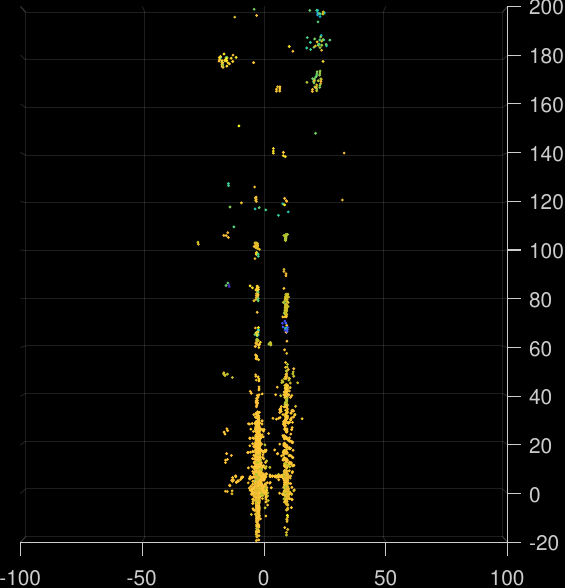} & \addpic{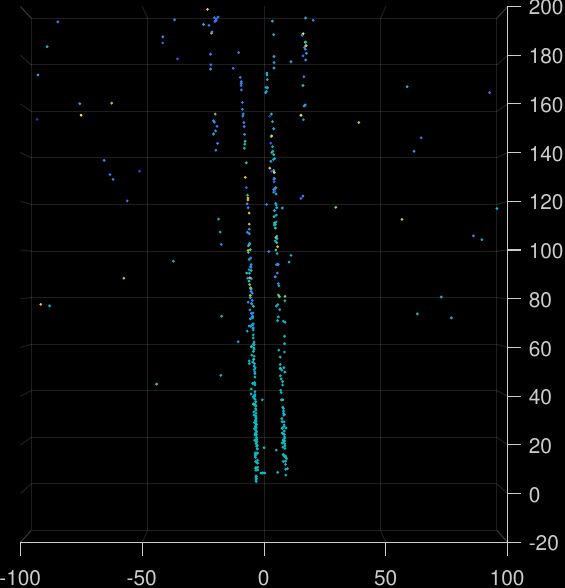} \\
\bottomrule
\end{tabular}
\caption{Samples of four sensors in different scenarios. D: Day, R: Rain, N: Night, H: Highway, T: Tunnel.
Rain causes many speckles in the lidar point cloud. 
The enhanced point cloud of Oculii Eagle formed by accumulating multiple frames has more points than a single frame of ARS548.
The accumulation also leads to some points behind the platform.
The tunnel data's perceptual aliasing causes the lidar(-inertial) odometry to give wrong velocity disagreeing with those from the GNSS/INS system and the two radars.
}
\label{tab:samples}
\end{table*}

\begin{figure}[!h]
	\centering
	\includegraphics[width=0.95\columnwidth]{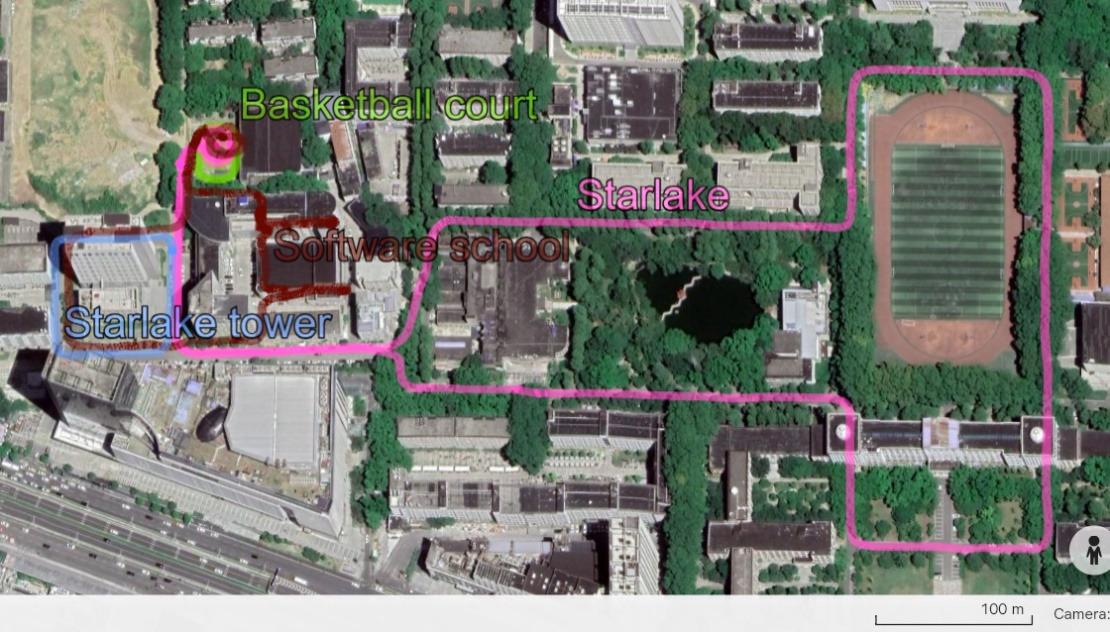} \\
	\vspace{2pt}
	\includegraphics[width=0.95\columnwidth]{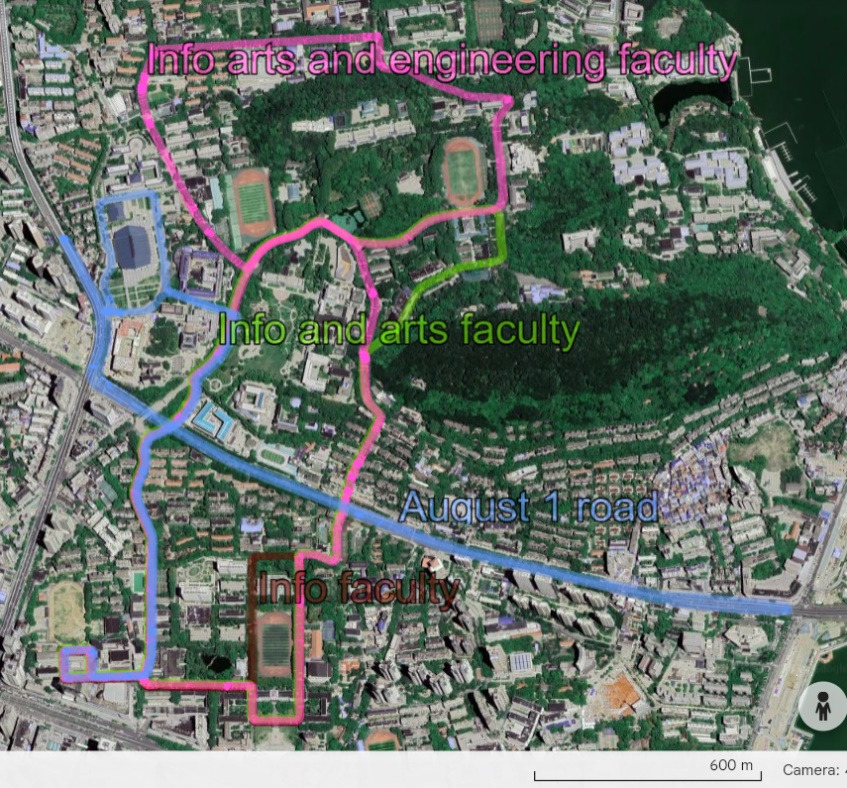}
	\caption{The routes used for data collection in our dataset. The top plot shows four routes: the basketball court (yellow), the starlake (pink), the software school (brown), and the starlake tower (blue).
	The bottom plot shows the other four routes: the info arts and engineering faculty (pink), the info and arts faculty (yellow),
	the August 1 road (blue), and the info faculty (brown). We slightly offset the paths for clarity.}
	\label{fig:places}
\end{figure}

There are a total of eight routes, as illustrated in Fig.~\ref{fig:places}:
the basketball court, the starlake, the software school, the starlake tower, the info faculty, 
the info and arts faculty, the info arts and engineering faculty, and the August 1 road.
The basketball court is a small flat area surrounded by some buildings.
The starlake route traverses the campus, featuring dense vegetation.
The software school route passes through urban canyons formed by tall buildings.
The starlake tower route encircles the high-rise starlake tower.
Since the Wuhan University main campus consists of three segments, the info faculty, the arts and sciences faculty, and the engineering faculty, we collected data on three incremental routes spanning these segments, featuring urban roads in the campus.
The August 1 road route is a highway including a long tunnel.

\subsection{Sensor setups}

\begin{figure*}[!h]
	\centering
	\includegraphics[width=0.372\columnwidth]{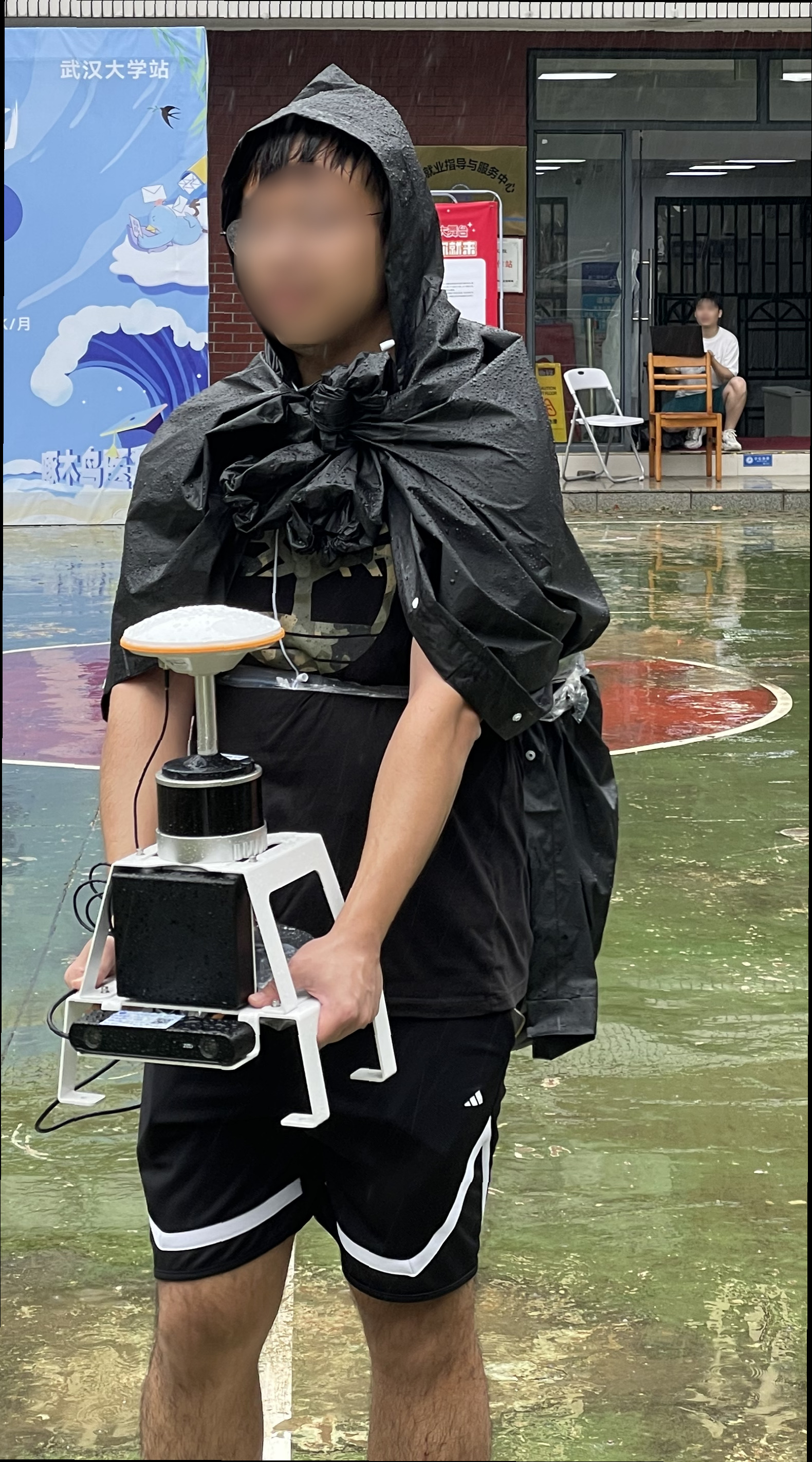}
	\includegraphics[width=0.85\columnwidth]{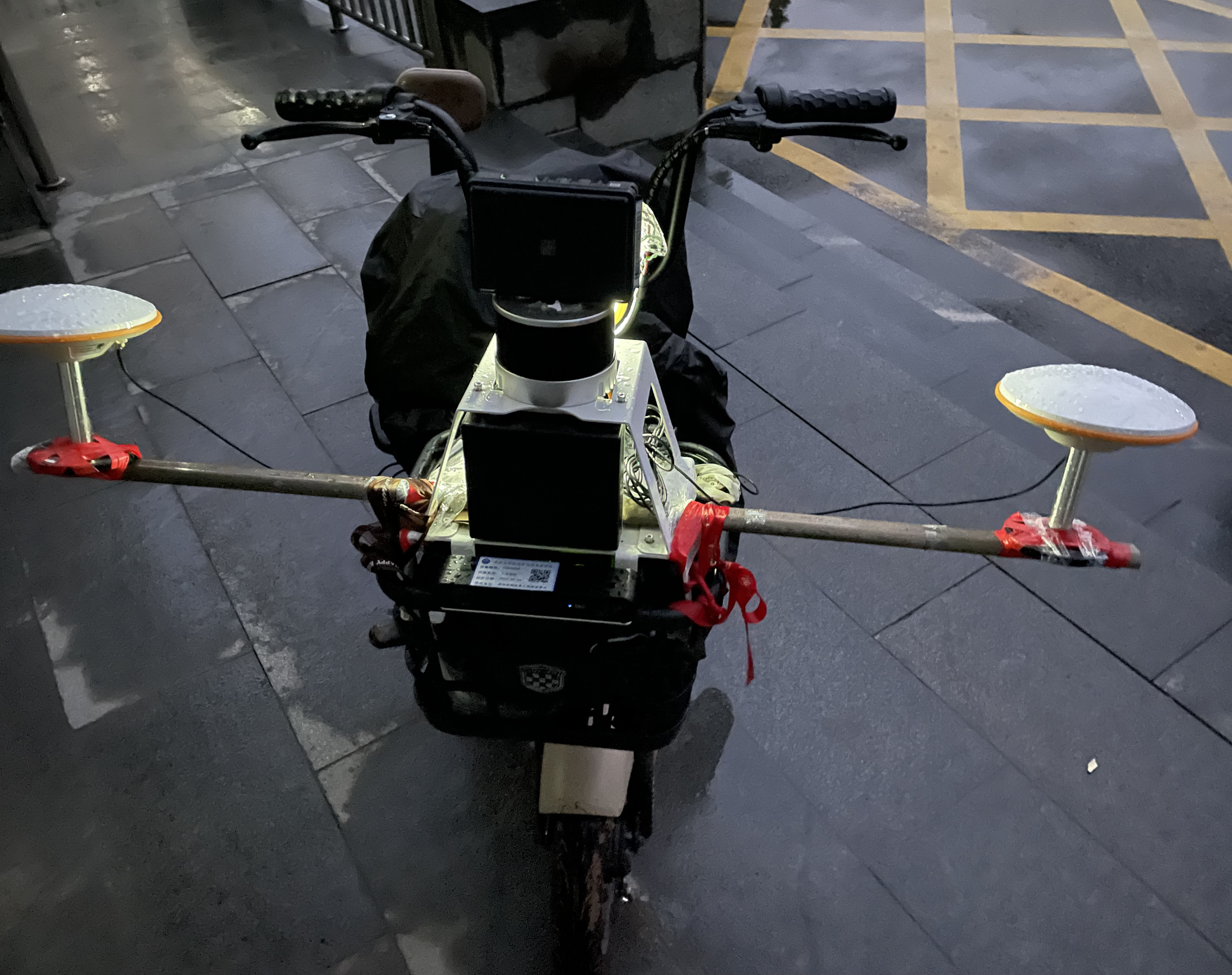}
	\includegraphics[width=0.61\columnwidth]{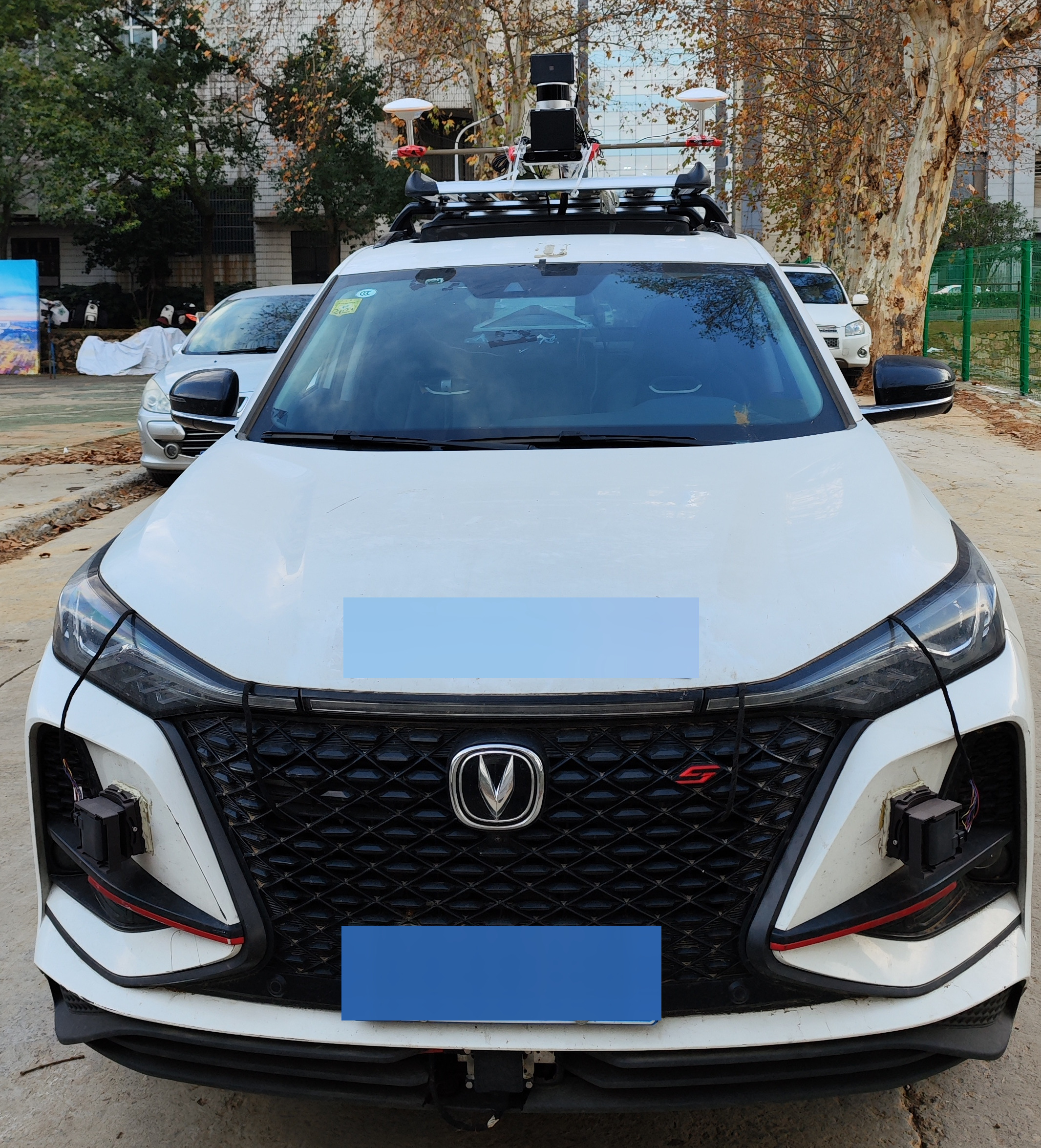} \\
(a)\hspace{0.61\columnwidth}(b)\hspace{0.73\columnwidth}(c)\\
	\includegraphics[width=0.95\columnwidth]{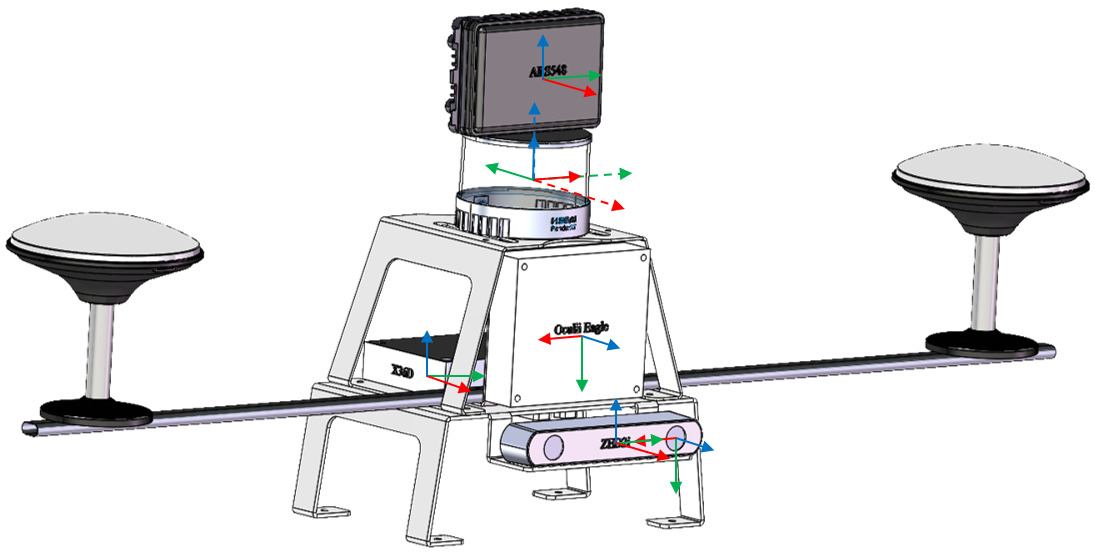}
	\includegraphics[width=0.9\columnwidth]{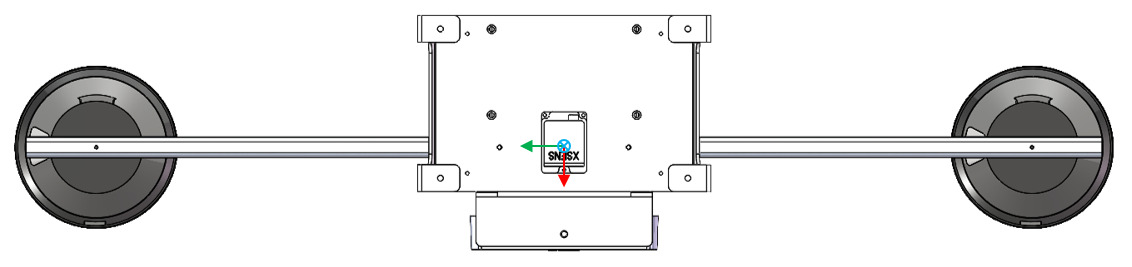} \\
(d)\hspace{0.92\columnwidth}(e)
	\caption{The sensor rig on three platforms and the coordinate frames of the sensors.
	The Thinkpad P53 laptop and the bundle of cables are concealed in a waterproof bag.}
	\label{fig:sensorrig}
\end{figure*}

We assembled the sensors on an aluminum alloy frame designed with CAD, while ensuring it was waterproof.
The sensor setups for three platforms are drawn in Fig.~\ref{fig:sensorrig}, with sensors for data collection listed in Table~\ref{tab:sensors}.
The handheld rig consists of a Hesai Pandar XT32 lidar, an Oculii Eagle radar, a ZED2i stereo camera, and a Bynav X36D GNSS/INS system with a single antenna.
The rig mounted on the ebike or the SUV includes the Hesai lidar, the Oculii radar, the ZED2i camera, the Bynav X36D GNSS/INS system with dual antennas,
and two additional sensors, a Continental ARS548 radar and an XSens MTi3DK IMU.
Specifically, sequences captured before Nov 1 2023 do not include data from the ARS548 radar and MTi3DK IMU.
These sequences encompass all handheld sequences and some ebike sequences.

\begin{table}[]
	\centering
	\caption{The descriptions of sensors in our dataset. The accelerometer noise and random walk are in units of $m/s^2/\sqrt{Hz}$ and $m/s^3/\sqrt{Hz}$. The gyroscope noise and random walk are in units of $rad/s/\sqrt{Hz}$ and $rad/s^2/\sqrt{Hz}$. FOV: field of view, BS: bias stability.}
	\label{tab:sensors}
	\begin{tabular}{@{}c|ccl@{}}
		\toprule
		\textbf{Type} &
		\textbf{Sensor} &
		\textbf{Rate} &
		\multicolumn{1}{c}{\textbf{Characteristics}} \\ \midrule
		\multirow{2}{*}{\begin{tabular}[c]{@{}c@{}}4D\\ Radar\end{tabular}} &
		\begin{tabular}[c]{@{}c@{}}Conti\\ ARS548\end{tabular} &
		15 &
		\begin{tabular}[c]{@{}l@{}}max range 300 m\\ Doppler {[}-400, 200{]} m/s\\ HFOV 120$^\circ$ VFOV 40$^\circ$\\ range accuracy 0.3 m\\ az. accuracy 0.2$^\circ$\\ elev. accuracy 0.1$^\circ$\\ \#point/frame $\sim$150\end{tabular} \\ \cmidrule(l){2-4} 
		&
		\begin{tabular}[c]{@{}c@{}}Oculii\\ Eagle\end{tabular} &
		15 &
		\begin{tabular}[c]{@{}l@{}}max range 400 m\\ Doppler $\pm$86.8 m/s\\ HFOV 113$^\circ$ VFOV 45$^\circ$\\ range accuracy 0.86 m\\ az. accuracy 0.44$^\circ$\\ elev. accuracy 0.175$^\circ$\\ \#point/frame $\sim$7500\end{tabular} \\ \midrule
		\begin{tabular}[c]{@{}c@{}}GNSS/\\ INS\end{tabular} &
		\begin{tabular}[c]{@{}c@{}}Bynav\\ X36D\end{tabular} &
		100 &
		\begin{tabular}[c]{@{}l@{}}real-time GNSS/INS\\ H 0.235 m RMS\\ V 0.14m RMS\\ in 10 s outage\end{tabular} \\ \midrule
		\multirow{3}{*}{IMU} &
		\begin{tabular}[c]{@{}c@{}}ZED2i\\ IMU\end{tabular} &
		400 &
		\begin{tabular}[c]{@{}l@{}}accel. noise 1.6$\times10^{-3}$, \\ random walk 2.5$\times10^{-4}$;\\ gyro. noise 1.2$\times10^{-4}$,\\ random walk 3.4$\times10^{-5}$\end{tabular} \\ \cmidrule(l){2-4} 
		&
		MTi3DK &
		100 &
		\begin{tabular}[c]{@{}l@{}}roll/pitch 0.5$^\circ$ RMS\\ yaw 2$^\circ$ RMS\\ accel. noise 7$\times10^{-4}$,\\ BS 4$\times10^{-4} m/s^2$;\\ gyro. noise 5.236$\times10^{-5}$,\\ BS 6$^\circ$/hr\end{tabular} \\ \cmidrule(l){2-4} 
		&
		\begin{tabular}[c]{@{}c@{}}X36D\\ IMU\end{tabular} &
		100 &
		\begin{tabular}[c]{@{}l@{}}accel. noise 5.833$\times10^{-4}$,\\ BS 1.5$\times10^{-4} m/s^2$;\\ gyro noise 2.91$\times10^{-5}$,\\ BS 1.8$^\circ$/hr\end{tabular} \\ \midrule
		Lidar &
		\begin{tabular}[c]{@{}c@{}}Hesai\\ Pandar\\ XT32\end{tabular} &
		10 &
		\begin{tabular}[c]{@{}l@{}}max range 120 m\\ range accuracy $\pm$2 cm\\ FOV 360$^\circ\times$31$^\circ$\\ \#point/frame $\sim$90000\end{tabular} \\ \midrule
		\begin{tabular}[c]{@{}c@{}}Stereo\\ camera\end{tabular} &
		ZED2i &
		20 &
		\begin{tabular}[c]{@{}l@{}}1280$\times$720, grayscale,\\ HFOV $110^\circ$, VFOV $70^\circ$\end{tabular} \\ \bottomrule
	\end{tabular}
\end{table}

A ThinkPad P53 laptop running Ubuntu 20.04 with a 1TB solid-state drive is used for real-time data preprocessing and recording. 
This laptop supports GPU processing and is compatible with all used ROS drivers. 
It connects to the internet via a WiFi hotspot hosted by a mobile smartphone with 5G networking service.
The internet connection is primarily used for updating the computer time via NTP (Network Time Protocol) and 
receiving RTCM (Radio Technical Commission for Maritime Services) messages used in RTK GNSS positioning, broadcast by the Qianxun FindCM service.

The ARS548, Oculii, and XT32 are connected to the laptop by Ethernet cables, 
while the ZED2i and MTi3DK are connected via USB Type-C connectors.
The X36D connects to the computer through both a USB A connector and an RJ45 connector. 
The USB A connection (at COM1) is for receiving control commands, 
and the RJ45 connection is for sending INS solutions and raw data to the computer, 
as well as receiving RTCM messages relayed by the computer.
The X36D's PPS signals and GPGGA messages at COM1 are directly fed to the Hesai XT32 through a wired connection to its GPS port,
ensuring that the lidar messages are synchronized to the GNSS time.

For the XT32, Eagle, MTi3DK, and ZED2i, data is captured as-is by their official ROS drivers.
The ARS548 and X36D UDP (User Datagram Protocol) packets are recorded on-site using the \texttt{tshark} utility.

During daytime data collection, the ZED2i's auto-exposure is enabled, usually adjusting the exposure time to less than 5 ms, often around 2 ms.
At dusk or night, auto-exposure is disabled, and the exposure time is locked at 5 ms to reduce motion blur.

All data collections start from an open area.
Every time the laptop and all sensors are powered up, we first ensure that the GNSS solution is fixed, then move the platform around to perform INS alignment until the INS solution converges.

\subsection{File formats}
The sequences are organized into folders named by their respective dates, with each folder containing all sequences captured on that time of day.
For each sequence, both a monolithic zipped ROS bag file and a zipped folder containing individual message files are provided.
The dataset is available for download on Microsoft OneDrive and Baidu Netdisk.

The message types in a rosbag and their respective folders are listed in Table~\ref{tab:topics}.
In these folders, the point clouds from the Pandar XT32 lidar, the ARS548 radar, and the Eagle radar are saved as \texttt{pcd} files for easy visualization.
The compressed images of the ZED2i stereo camera are saved as \texttt{jpg} images.
The other topics, \eg, GNSS/INS solutions, and IMU data are saved in \texttt{txt} files.
The fields for these sensor messages are detailed in a series of CSV tables available on the dataset website.
To facilitate easy frame transformation, we include a ROS1 launch file in each sequence's folder.
This file publishes the static transforms between the available sensors in the sequence.

The ARS548 point clouds are decoded from the detection lists in UDP packets.
The X36D messages are also decoded from the UDP packets following the Bynav protocol 
\endnote{Bynav data communication interface protocol UG016  \url{https://www.bynav.com/cn/resource/support?cat=6}}.

\begin{table*}[]
	\centering
	\caption{The ROS1 topics and message types in the published rosbags, along with the corresponding folders in the published zip files.}
	\label{tab:topics}
	\begin{tabular}{@{}lcccc@{}}
		\toprule
		\multicolumn{1}{c}{\textbf{Sensor}} &
		\textbf{Topic} &
		\textbf{ROS message} &
		\textbf{Folder} &
		\textbf{Format} \\ \midrule
		Conti ARS548                & /ars548                   & PointCloud2 & ars548/points & pcd \\ \midrule
		Hesai XT32                  & /hesai/pandar             & PointCloud2 & xt32          & pcd \\ \midrule
		MTi3DK IMU                  & /mti3dk/imu               & Imu         & mti3dk        & txt \\ \midrule
		\multirow{3}*{Oculii Eagle} &
		/radar\_enhanced\_pcl2 &
		PointCloud2 &
		eagleg7/enhanced &
		pcd \\
		& /radar\_pcl2              & PointCloud2 & eagleg7/pcl   & pcd \\
		& /radar\_trk               & PointCloud  & eagleg7/trk   & pcd \\ \midrule
		\multirow{3}{*}{Bynav X36D} & /x36d/gnss                & NavSatFix   & x36d          & txt \\
		& /x36d/gnss\_ins           & NavSatFix   & x36d          & txt \\
		& /x36d/imu\_raw            & Imu         & x36d          & txt \\ \midrule
		\multirow{4}{*}[-1.5em]{ZED2i}      & /zed2i/zed\_node/imu/data & Imu         & zed2i         & txt \\
		&
		\begin{tabular}[c]{@{}c@{}}/zed2i/zed\_node/left\_raw/\\ image\_raw\_gray/compressed\end{tabular} &
		CompressedImage &
		zed2i/left &
		jpg \\
		&
		\begin{tabular}[c]{@{}c@{}}/zed2i/zed\_node/right\_raw/\\ image\_raw\_gray/compressed\end{tabular} &
		CompressedImage &
		zed2i/right &
		jpg \\
		& /zed2i/zed\_node/odom     & Odometry    & zed2i         & txt \\ \bottomrule
	\end{tabular}
\end{table*}

For each sequence, we provide reference trajectories of the lidar XT32 frame in the TLS map frame at 10 Hz.
However, for large SUV sequences, we only provide the reference poses during the starting subsequence and end subsequence which are covered by the TLS point clouds.
The full trajectories generated by the cascaded pose graph optimization (CPGO) \citep{huaiCollaborativeMonocularSLAM2018} are also provided, 
but their height accuracy is only within 1 meter.
These full trajectories are intended for place recognition instead of odometry benchmarking.
To be complete, we provide all the TLS point clouds and the refined poses of these TLS scans.

For all sequences, we also provide the real-time kinematic (RTK) GNSS/INS solutions,
\ie, the latitude, longitude, and ellipsoid height wrt the WGS84 (EPSG4326) reference frame,
 the used GNSS service, the solution status, 
and diagonal position covariance entries extracted from the Bynav messages \texttt{INSPVAXA} and \texttt{RAWIMUSA}.
The coordinates are also converted to the UTM50N zone (EPSG 32650), since Wuhan is located at the northwest corner of UTM50R zone.
The INS solution is sometimes unreliable in the vertical direction, having spikes up to a few meters.

A python-based SDK is also provided to load the sensor data from the sequence folders and to visualize. Also, the SDK supports conversion between the ROS1 bags and sequence folders.
The sensor calibration data are provided in MATLAB scripts \endnote{Dataset tools \url{https://github.com/snail-radar/dataset_tools}}
and ROS1 launch files with \texttt{static\_transform\_publisher}s.

\section{Reference trajectories}
\label{sec:truth}
The GNSS RTK/INS solutions from the Bynav system often exhibit height jumps of up to a few meters due to the frequent GNSS outages \citep{tothRemoteSensingPlatforms2016}.
Therefore, we generate the reference poses using precise TLS point clouds, similar to  \citep{ramezaniNewerCollegeDataset2020}.
But instead of frame-wise ICP, we sequentially align the undistorted lidar frames to the TLS map using a LIO method.
A limitation of these reference poses is that they are only available at the beginning and the end of those long sequences.
Despite the limitation, proper odometry evaluation is still feasible \citep{zhangTutorialQuantitativeTrajectory2018}.
For completeness, we also provide the full reference trajectories, 
generated using CPGO.

\subsection{TLS-based lidar inertial localization}
A total of 93 TLS scans were captured by a Leica RTC360 scanner on a sunny day, covering the starlake and the starlake tower routes.
In capturing these scans, we ensured sufficient overlap between consecutive scans, with
a mean distance of 28.2 m between consecutive scans.
These scans were first processed using the Cyclone Register 360 program, with manually identified additional loop constraints between some scans.
The TLS scan registration results were further refined by point-to-plane ICP in Open3D
and regularized by SE(3) PGO with uniform weights while considering the two loops within these scans.
The pairwise registration results were inspected by three individuals over approximately 10 rounds to ensure proper pairwise matching.
The position accuracy of the TLS pairwise registration is expected to be within 5 cm, as implied by the quality check in Table~\ref{tab:ref-traj-stats}.
The final TLS map was created by stitching together these 93 scans.

Next, for each sequence, we generate the reference trajectory relative to the TLS map in three steps like the classic Rauch-Tung-Striebel Smoother.
First, we determine the initial pose, then perform LIO in both forward and backward localization modes, 
and finally average the results from forward and backward localization to produce the reference trajectory.

To get the initial pose, we selected a lidar scan at the stationary beginning, and registered the scan to a nearby TLS scan 
in the CloudCompare program by manually picking five pairs of correspondences.
The labor was acceptable as our routes used fewer than five unique starting poses separated by a notable distance.
Every initial pose was further refined with the classic point-to-plane ICP in Open3D and manually checked in terms of the number of matches and inlier RMSE (root mean squared error).
With these initial poses, we ran an extended FAST-LIO2 method \citep{xuFASTLIO2FastDirect2022} in localization mode.
This method loads the constant TLS map and predicts the pose with IMU data,
undistorts the lidar scan, and iteratively refines the pose with matches between scan points and map surfels.
The LIO method operates in localization mode as the incremental submap building is replaced by loading segments of the TLS map in a sliding manner.
Also, to remove the real-time operation constraint and execution randomness, the method was converted to run sequentially in offline mode.
The IMU data used in the localization method is from the MTi3DK or the X36D when the MTi3DK is unavailable.

For large-scale sequences, since the TLS map only covers the beginning and end parts of the sequence,
we ran the localizer only for the beginning and end subsequences within the TLS coverage, as done in TUM-VI \citep{schubertTUMVIBenchmark2018}.
We selected data from the first four minutes as the start subsequence and data from the last four minutes as the end subsequence.
Each subsequence was processed by the LIO in localization mode which was terminated when the platform's minimum distance to the TLS trajectories exceeded 8 m.
Here the TLS trajectories were simply the reference poses of multiple medium-scale sequences which covered the entire TLS area.

We also ran the localizer in backward mode to get the backward trajectory, where both the beginning and end subsequences were reversed.
The reverse sequence was created with the requirement that the platform poses remained unchanged in space despite being reversed in time.
Specifically, we first found the maximum time of the sequence $t_{max}$, then modified
every message's stamp and every lidar point's stamp to $t^\prime = 2t_{max} - t$ from its original time $t$.
The accelerometer data remained intact $\mbf a_m^\prime \leftarrow \mbf a_m$,
but the gyro data were inverted in sign $\boldsymbol \omega_m^\prime \leftarrow -\boldsymbol{\omega}_m$.
After the processing with LIO in localization mode,
we removed the last 8 seconds of the resulting poses for the forward pass of the beginning subsequence and the backward pass of the end subsequence, 
as these poses were inaccurate.
The pose times of the backward pass were then restored using $t_{max}$.

Finally, with the forward and backward trajectories, we took their averages to get the reference trajectory.

The backward pass offers a means to validate the reference trajectory accuracy
by comparing the forward and backward results against their averages.
For each sequence, we computed the position and rotation difference magnitudes (denoted by $\delta \mathbf{p}_{WL}$ and $\delta \mathbf{R}_{WL}$, respectively)
between the forward pass poses and the reference poses.
The median and maximum values of these differences are listed in Table~\ref{tab:ref-traj-stats}.
Except one trajectory, all sequences have median deviations less than 5 cm and 0.5$^\circ$.
Larger deviations typically occur during turns and on bumpy roads.

\begin{figure*}[htbp]
	\centering
	\includegraphics[width=0.95\columnwidth]{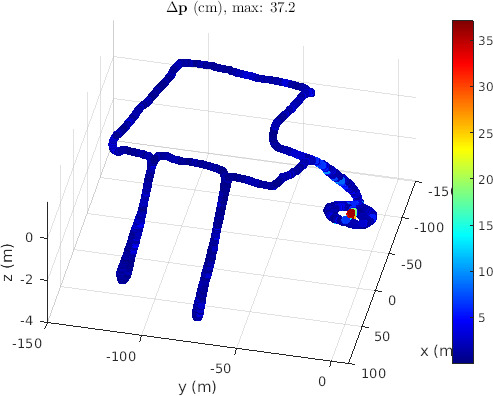}
	\includegraphics[width=0.95\columnwidth]{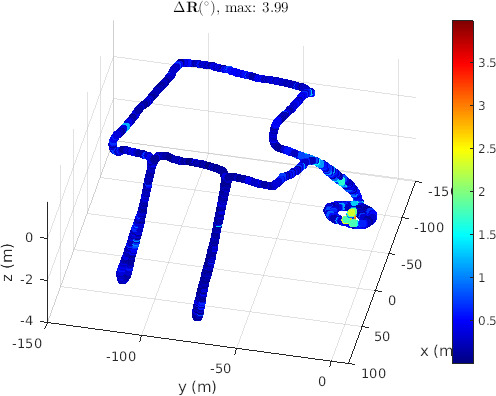}\\
	\qquad \qquad (a) \qquad \qquad (b) \\
	\includegraphics[width=0.95\columnwidth]{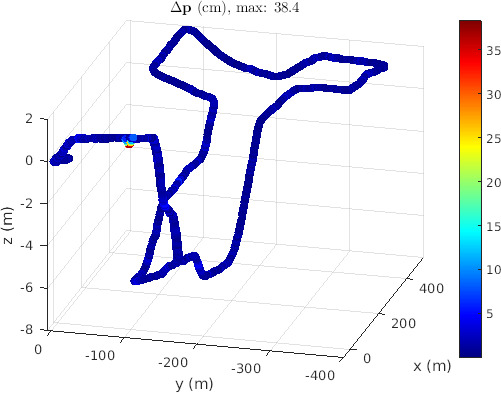}
	\includegraphics[width=0.95\columnwidth]{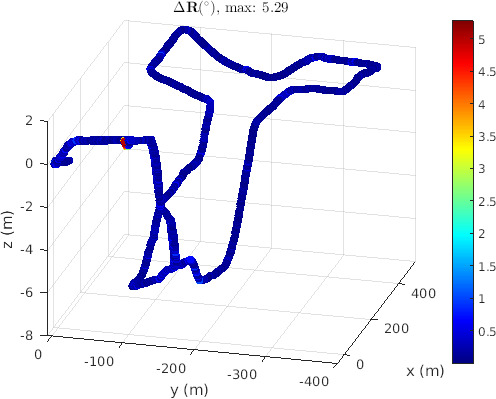}\\
	\qquad \qquad (c) \qquad \qquad (d)\\
\caption{
The translation (a, c) and rotation (b, d) differences between the forward (or backward) localization and their averages are 
shown for the e-bike sequences 20231105/4 (a, b) and 20231105\_aft/4 (c, d).
These two sequences exhibit the largest maximum translation deviations among all sequences in the dataset.
}
\label{fig:fb-diff}
\end{figure*}

For the 20231105/4 and 20231105\_aft/4 sequences, we plot the position and rotation deviations along the reference trajectory using a jet colormap in Fig.~\ref{fig:fb-diff}. 
The results show that large deviations are confined to specific locations.

\begin{table}[]
	\centering
	\caption{The median and maximum differences in translation and rotation between the forward (or backward) trajectory and their averages.}
	\label{tab:ref-traj-stats}
	\begin{tabular}{cccccc}
		\hline
		\multicolumn{1}{c|}{\multirow{2}{*}{Date/Run}} &
		\multicolumn{1}{c|}{\multirow{2}{*}{Dist(m)}} &
		\multicolumn{2}{c|}{$\Delta \mathbf{p}$(cm)} &
		\multicolumn{2}{c}{$\Delta \mathbf{R}(^\circ)$} \\ \cline{3-6} 
		\multicolumn{1}{c|}{} &
		\multicolumn{1}{c|}{} &
		\multicolumn{1}{c|}{med.} &
		\multicolumn{1}{c|}{max} &
		\multicolumn{1}{c|}{med.} &
		max \\ \hline
		20230920/1      & 82.3   & 0.8 & 4.5  & 0.11 & 3.86 \\
		20230920/2      & 2331.8 & 1.2 & 6.3  & 0.40 & 2.61 \\
		20230921/2      & 53.5   & 1.0 & 4.9  & 0.11 & 0.39 \\
		20230921/3      & 2043.6 & 1.2 & 9.4  & 0.17 & 0.82 \\
		20230921/4      & 716.4  & 0.9 & 9.2  & 0.17 & 1.83 \\
		20230921/5      & 1976.6 & 1.0 & 11.0 & 0.14 & 0.75 \\
		20231007/2      & 1983.5 & 0.9 & 6.7  & 0.14 & 1.78 \\
		20231007/4      & 232.3  & 0.9 & 11.4 & 0.14 & 1.35 \\
		20231019/1      & 1916.6 & 1.0 & 12.1 & 0.17 & 2.02 \\
		20231019/2      & 750.9  & 0.7 & 9.4  & 0.15 & 1.66 \\
		20231105/2      & 2004.3 & 1.1 & 32.9 & 0.16 & 1.95 \\
		20231105/3      & 2062.8 & 1.1 & 15.0 & 0.14 & 4.70 \\
		20231105/4      & 897.7  & 1.4 & 37.2 & 0.28 & 3.99 \\
		20231105/5      & 946.4  & 1.3 & 6.9  & 0.23 & 1.92 \\
		20231105/6      & 250.1  & 0.8 & 7.2  & 0.13 & 1.45 \\
		20231105\_aft/2 & 166.2  & 0.7 & 2.0  & 0.09 & 0.39 \\
		20231105\_aft/4 & 2000.8 & 1.2 & 38.4 & 0.17 & 5.29 \\
		20231105\_aft/5 & 810.8  & 0.8 & 10.5 & 0.15 & 1.31 \\
		20231109/3      & 1977.8 & 1.1 & 13.9 & 0.12 & 1.57 \\
		20231109/4      & 729.4  & 0.5 & 4.5  & 0.05 & 1.38 \\
		20231201/2      & 1737.3 & 1.2 & 10.1 & 0.04 & 0.24 \\
		20231201/3      & 1724.3 & 1.2 & 7.2  & 0.04 & 0.42 \\
		20231208/1      & 275.3  & 0.6 & 6.5  & 0.02 & 0.19 \\
		20231208/4      & 1718.7 & 1.2 & 6.5  & 0.03 & 0.20 \\
		20231208/5      & 1731.7 & 1.3 & 8.1  & 0.04 & 0.29 \\
		20231213/1      & 276.1  & 1.1 & 7.0  & 0.03 & 0.22 \\
		20231213/2      & 1730.6 & 1.3 & 9.3  & 0.04 & 0.51 \\
		20231213/3      & 1706.7 & 1.4 & 10.3 & 0.04 & 0.41 \\
		20231213/4      & 1769.8 & 1.4 & 9.0  & 0.04 & 0.36 \\
		20231213/5      & 1755.9 & 1.4 & 9.0  & 0.03 & 0.25 \\
		20240113/1      & 541.4  & 1.0 & 10.4 & 0.03 & 3.62 \\
		20240113/2      & 1528.1 & 1.1 & 7.2  & 0.03 & 0.21 \\
		20240113/3      & 1736.9 & 1.3 & 6.9  & 0.03 & 0.54 \\
		20240113/5      & 1728.1 & 1.2 & 6.8  & 0.03 & 0.20 \\
		20240115/2      & 1742.2 & 1.3 & 8.9  & 0.03 & 0.19 \\
		20240115/3      & 1771.9 & 1.3 & 8.7  & 0.03 & 0.20 \\
		20240116/2      & 648.2  & 1.1 & 7.5  & 0.04 & 0.24 \\
		20240116/4      & 1706.9 & 1.4 & 8.7  & 0.03 & 0.20 \\
		20240116/5      & 1744.8 & 1.2 & 7.7  & 0.03 & 1.74 \\
		20240116\_eve/3 & 659.2  & 1.2 & 6.6  & 0.03 & 0.23 \\
		20240116\_eve/4 & 1715.0 & 1.3 & 7.1  & 0.03 & 0.21 \\
		20240116\_eve/5 & 1737.7 & 1.4 & 7.2  & 0.04 & 0.36 \\
		20240123/2      & 644.6  & 1.3 & 7.4  & 0.04 & 0.20 \\
		20240123/3      & 1761.9 & 1.2 & 8.4  & 0.03 & 0.23 \\ \hline
	\end{tabular}
\end{table}

\subsection{Cascaded pose graph optimization}
To generate full trajectories for sequences extending beyond the TLS map, CPGO \citep{huaiCollaborativeMonocularSLAM2018} is applied
to fuse multi-source constraints.
These constraints include relative poses from the KISS-ICP odometry \citep{vizzo2023ral}, 
absolute poses from TLS-based lidar-inertial localization,
absolute positions from the GNSS/INS system, 
and preintegrated inertial factors from the X36D IMU.

State variables in CPGO consist of the sequential poses and velocities of the XT32 lidar in the TLS map frame,
sequential IMU biases, and the gravity direction in the TLS map frame.
At first, we solve for the transformation between the UTM50N frame (used for GNSS/INS data) and the TLS map frame 
by treating it as a variable in CPGO, using a long sequence fully covered by the TLS map (20240921/data5).
For other sequences, this transformation is kept constant during CPGO.

CPGO proceeds in three steps.
First, it only optimizes the rotation components of the sequential poses
by using the relative odometry and absolute TLS poses.
Second, it only optimizes the translation components of the sequential poses 
by using the same constraints.
Finally, CPGO refines all state variables by using all available constraints.
In this last step, we used the Cauchy loss for relative odometry constraints and 
GNSS/INS absolute positions to deal with lidar odometry drift in tunnels and GNSS outages.
A set of weights for these constraints were manually tuned to improve trajectory agreement across repetitive sessions and used for all sequences.
Empirically, we found that cascaded optimization enlarged the convergence basin of the PGO problem.
The resulting full trajectories for the XT32 lidar are provided relative to the UTM50N frame.
By checking the deviation between two challenging sessions of the August 1 road, 
we see that the horizontal accuracy of the PGO trajectory is better than 0.15 m, and the vertical accuracy is within 1 meter.

\section{Synchronization}
\label{sec:sync}
Since it is challenging to sync all sensors in hardware, we propose a scheme to ensure that their messages are stamped by the same virtual clock.
The scheme first uses the lidar data as a bridge to map all sensor message timestamps to the GNSS time.
Then, the constant time offset between message topics is estimated by odometry and correlation algorithms.

We first discuss the lidar-based synchronization.
Messages from all sensors except the X36D GNSS/INS system have both a local timestamp from the host/laptop clock and a remote timestamp from the sensor clock.
For a message topic, to remove the jitter in the host times due to transmission, 
we use the convex hull algorithm \citep{zhangClockSynchronizationAlgorithms2002,rehderGeneralApproachSpatiotemporal2016} for smoothing,
which gives the smooth host time for a message on the topic given the message's sensor time and host time, \ie,
a mapping function $s_{topic}(\cdot, \cdot):\texttt{sensor time} \times \texttt{host time}\rightarrow \texttt{smooth host time}$ for messages on the topic.
Moreover, the XT32 lidar is synced to the GNSS time thanks to the X36D GNSS/INS system.
That is, we can get the GNSS time given the smooth host time of a lidar message, 
which is a mapping $g_{xt32}(\cdot): \texttt{smooth host time}\rightarrow \texttt{GNSS time}$.
Since host times of messages of all topics are stamped by the laptop, we can obtain the GNSS time of a message using
$g_{xt32}(s_{topic}(\texttt{sensor time}, \texttt{host time}))$ which removes the jitter and long-term clock drift.
An advantage of this sync step is that it can be executed causally.

After mapping to the GNSS time, a small constant time offset may persist between two streams of messages, \eg, the ZED2i IMU and the ZED2i images. due to transmission delays.
This constant time offset between two motion-related message topics can be found using the correlation method.
Since all messages in our dataset are motion-related except the camera information messages,
we estimate the time offset of these message streams relative to the X36D GNSS/INS trajectory.

Specifically, for the XT32 lidar, we first processed the lidar scans with the lidar odometry (LO) method, KISS-ICP \citep{vizzo2023ral},
which gave 3D lidar poses at the middle of each scan.
Next, the angular rates in the lidar frame \{L\} and the INS frame \{I\} were computed from the INS poses and LO poses by central difference, obtaining $\boldsymbol{\omega}_{WL}^L$ and $\boldsymbol{\omega}_{WI}^I$.
Finally, the time delay of the lidar data was computed by the Correlation Algorithm \ref{alg:correlation}, given the angular rates and the nominal $\mathbf{R}_{IL}$.

For a 4D radar, \eg, the ARS548 (topic \texttt{ars548}) or the Oculii Eagle (topic \texttt{radar\_pcl2}), we first processed the radar scans with the GNC method \citep{yangGraduatedNonconvexityRobust2020} to estimate its ego velocity, $\mathbf{v}_{WR}^R$.
Next, the ego velocity $\mathbf{v}_{WR}^I$ was computed from the INS poses with the nominal lever arm $\mathbf{p}_{IR}^I$ by central difference.
Finally, the time delay of the radar data was estimated using the correlation algorithm, given the ego velocities and the nominal $\mathbf{R}_{IR}$.

For the IMUs, \eg, the ZED2i IMU, which directly provides the angular velocity, $\boldsymbol{\omega}_{WU}^U$ in its frame \{U\},
we first computed the angular rates in the INS frame, $\boldsymbol{\omega}_{WI}^I$, by central differencing the INS poses,
and then ran the correlation algorithm given the angular rates and the nominal $\mathbf{R}_{IU}$.

The correlation algorithm \ref{alg:correlation} finds the time offset $t_d$ and relative rotation $\mathbf{R}_{AB}$ 
between two stamped sequences $\{\mathbf{v}_i^A\}$ at $\{s_i, i=1,\cdots, N_a\}$ and $\{\mathbf{v}^B_j\}$ at $\{t_j, j=1,\cdots, N_b\}$, \eg, angular rates or ego velocities.
Its idea to find the maximum cross correlation is straightforward,
but there are a few parameters worth noting.
The default smoothing window sizes $M_a$ and $M_b$ are 1, 
meaning no smoothing is applied. 
When the two sequences have very different rates, \eg, 100 vs 10, a smoothing window of size \eg, 5, is necessary for the higher rate sequence.
This smoothing causes signal delay which should be accounted for when computing the time offset $t_d$.
The time tolerance for associating the time-corrected data should typically be smaller than a fraction of the resampling interval, \eg, $\delta t/2$.
The truncated least squares method in each iteration first
solves for the rotation perturbation $\delta\boldsymbol{\theta}$ and secondly update $\mathbf{R}_{AB}$
by $\mathbf{R}_{AB} \leftarrow \Exp{(\lfloor \delta\boldsymbol{\theta} \rfloor_\times)}\mathbf{R}_{AB}$.
$\delta\boldsymbol{\theta}$ is obtained by solving the linear system stacked from the basic residual equation,
\begin{equation}
\lfloor \mathbf{R}_{AB} \mathbf{v}^B\rfloor_\times \delta\boldsymbol{\theta} 
\doteq (\mathbf{R}_{AB} \mathbf{v}^B) \times \delta\boldsymbol{\theta} 
\approx \mathbf{v}^A - \mathbf{R}_{AB} \mathbf{v}^B,
\end{equation}
while discounting those with large residuals.

\begin{algorithm}
	\caption{Correlation Algorithm for 3D Motion Data}
	\label{alg:correlation}
	\begin{algorithmic}[1]
		\Require Two sequences of timestamped velocity data $\{\mathbf{v}_i^A\}$ at $\{s_i, i=1,\cdots, N_a\}$ and $\{\mathbf{v}^B_j\}$ at $\{t_j, j=1,\cdots, N_b\}$;
		rough guess of the rotation between the two global frames, $\mathbf{R}_{AB}$.
		\Ensure Estimated time offset between two sequences $t_d$ and refined $\mathbf{R}_{AB}$.

		\State \textbf{Keep} only the overlapping part (including a buffer at both ends) of the two sequences.
		\State \textbf{Smooth} the data by a sliding window of Gaussian weights to remove the high frequency components and improve correlation.
		Denote the window sizes by $M_a$ and $M_b$ for the two sequences.
		\State \textbf{Determine} the sampling interval $\delta t$ as the smaller mean interval of the two sequences, and allocate the sample times for both sequences, $\{s_i^\prime, i=1,\cdots, N_a^\prime\}$, and $\{t_j^\prime, j=1,\cdots, N_b^\prime\}$.
		\State \textbf{Cubic fit} the two sequences of data and get samples at the proposed sample times.
		\State \textbf{Compute} norms of the two sequences of data and cross-correlate the two norm sequences.
		\State \textbf{Locate} the peak correlation step, and refine its position $d$ by quadratic fitting.
		\State Compute the time offset $t_d = d \delta t + s_1^\prime - t_1^\prime + (M_b - M_a) \delta t / 2$.
		\State \textbf{Correct} the timestamps of $\{\mathbf{v}_j^B\}$ by adding $t_d$.
		\State \textbf{Associate} $\{\mathbf{v}_i^A\}$ and $\{\mathbf{v}_j^B\}$ in time with a tolerance $\tau$.
		\State \textbf{Refine} $\mathbf{R}_{AB}$ by the truncated least squares method.
	\end{algorithmic}
\end{algorithm}

Though the correlation algorithm is robust and globally optimal, it can still fail in certain cases, \eg, when multiple correlation peaks occur.
Failed cases were spotted by first checking whether the time offsets along 4 axes (norm, x, y, and z) are consistent and
whether the resulting relative rotation is close to the nominal value, 
and secondly visually checking the time alignment plots.
These failures, about 10 out of 300 correlation attempts, were manually corrected case-by-case by tweaking the smoothing window size and/or masking the first peak.

For the ZED2i, we estimated the time offsets between its IMU and two cameras using the Swift-VIO method \citep{huaiObservabilityAnalysisKeyframebased2022}.
During estimation, we fixed the rolling shutter readout time to 14.4 ms as reported in \cite{huaiAutomatedRollingShutter2023}.
Though the method often gave drifty trajectories on challenging sequences, the time offsets always quickly stabilized with tapering 3$\sigma$ bounds due to its strong observability.
We used the medians in the last 7 seconds as the time offset estimates.

Overall, our sync precision is expected to be better than 5 ms, considering the following facts.
1. The GNSS/INS solution has a rate of 100 Hz, and the correlation algorithm uses quadratic fitting to refine the maximum correlation position.
2. The reported $\sigma$ of Swift-VIO toward the end of a sequence is always less than 1 ms.
3. The exposure time of ZED2i is adjusted to be no more than 5 ms.

The sequences in the published dataset already have the message timestamps compensated by the estimated time offsets to the GNSS time.
Specifically, the stamps of all motion-related messages are corrected by the corresponding time offsets.
For points in every XT32 lidar frame, their timestamps are also corrected by the lidar time offset.

\section{Sensor rig calibration}
\label{sec:calib}
For the dataset, we define the body frame to have the same origin as the XT32 lidar,
but with an orientation such that its x-axis points forward, y-axis points left, and z-axis points up. Thus,
the body frame relates to the XT32 frame by a known rotation.
The extrinsic parameters of all sensors are provided relative to the body frame.

Initial relative positions between sensors are measured manually, and initial relative orientations are obtained from the CAD drawing \ref{fig:sensorrig}.
These parameters are then refined through several methods.

The extrinsic parameters of the ZED2i IMU or the MTi3DK relative to the XT32 lidar are refined using the lidar IMU calibration tool \citep{zhu2022robust}.
For our specific ZED2i, there is an open data sheet providing the camera intrinsic parameters, IMU noise parameters, and camera extrinsic parameters, downloadable from its website.
This allows us to relate these ZED2i sensors to the body frame by concatenating these extrinsic transforms.

For the Bynav system, we obtain the refined relative rotation between the XT32 lidar and the Bynav IMU by the correlation algorithm \ref{alg:correlation}.

The relative orientation of the Oculii Eagle and ARS548 radars to the X36D IMU is also refined using the correlation algorithm.
For the ARS548 radar mounted on top of the rig, it slightly tilted backward for the few sequences captured on Nov 5 2023.
Therefore, we provide its relative orientation for every sequence on that day.

\section{Known issues}
\label{sec:issues}
We have identified several issues in the dataset.
\begin{itemize}
	\item To save storage space, the resolution of the ZED2i stereo images has been reduced to 640$\times$360 in resolution since Dec 8 2023.
	\item In some sequences, the ZED2i images and IMU data have lower rates than the nominal values, likely due to low power voltage.
\end{itemize}

\section{Comparative studies}
\label{sec:comparison}
Since this dataset is intended for SLAM research, 
we show its use by evaluating several recent radar-based odometry and place recognition methods, 
validating its suitability as a benchmarking resource for radar-based SLAM techniques.

\subsection{Radar odometry}
Recently, numerous radar odometry methods have been proposed. For spinning radars, notable examples include HERO \citep{burnettRadarOdometryCombining2021} and CFEAR \citep{adolfssonLidarlevelLocalizationRadar2023}.
For single-chip radars, approaches include 
radar-only methods such as 4DRadarSLAM \citep{zhang4DRadarSLAM4DImaging2023} and DeepEgo \citep{zhuDeepEgoDeepInstantaneous2023};
radar-inertial methods using Doppler data, such as EKF-RIO \citep{doerRadarInertialOdometry2020} and DRIO \citep{chenDRIORobustRadarinertial2023};
radar-inertial methods leveraging point matching, such as 4D iRIOM \citep{zhuang4DIRIOM4D2023} and \citep{michalczykTightlycoupledEKFbasedRadarinertial2022};
and learning-based radar-inertial odometry methods like MilliEgo \citep{luMilliEgoSinglechipMmWave2020}.

Among them, we selected three recent radar-based odometry methods: 4DRadarSLAM \citep{zhang4DRadarSLAM4DImaging2023}, 
EKF-RIO \citep{doerRadarInertialOdometry2020}, and 4D iRIOM \citep{zhuang4DIRIOM4D2023}.
These methods were tested on all sequences using data from either the Oculii Eagle or the Continental ARS548 radar.
4D RadarSLAM, specifically designed to process Oculii's enhanced accumulative point clouds, struggled with
the much sparser ARS548 point clouds and the unaccumulated Oculii Eagle data, often failing to find matches.
As a result, we evaluated 4DRadarSLAM using the Oculii accumulated point clouds, while the other methods processed the original sparse point clouds.

We measured odometry accuracy using the Absolute Trajectory Error (ATE) metric with the trajectory evaluation tool of \cite{zhangTutorialQuantitativeTrajectory2018}.
The ATE root mean square errors (RMSEs) are presented in Table~\ref{tab:rio}.
Among the methods, 4D iRIOM outperformed the other two across most sequences.
However, the recurring large errors underscore that current radar odometry methods are far from robust.
While they perform well on select sequences, many sequences exhibit significant drift, indicating substantial room for improvement.

\begin{table}[]
	\centering
	\caption{Absolute trajectory error (ATE) in meters RMSE (root mean squared error) of radar-based odometry methods, including the 4D RadarSLAM (RS), EKF-RIO (ER), and 4D iRIOM without loop closure (IR), on the Oculii Eagle and Continental ARS548 4D radar data of the proposed dataset.
	The rows are grouped according to the routes as in Table~\ref{tab:seqs}.
	Note that the ARS548 data are unavailable for sequences before 20231105.
The 4D RadarSLAM needs to run with the enhanced accumulative Oculii radar data as it requires dense points for scan matching.
The other two methods, EKF-RIO and 4D iRIOM processes the single frame data from Eagle and ARS548 without accumulation for better results.
A '-' indicates that ARS548 data is not available.
}
	\label{tab:rio}
	\begin{tabular}{cccccc}
		\hline
		\multicolumn{1}{c|}{\multirow{2}{*}{Sequence}} & \multicolumn{3}{c|}{Eagle}                                                  & \multicolumn{2}{c}{ARS548}     \\ \cline{2-6} 
		\multicolumn{1}{c|}{}                          & \multicolumn{1}{c|}{RS} & \multicolumn{1}{c|}{ER} & \multicolumn{1}{c|}{IR} & \multicolumn{1}{c|}{ER} & IR   \\ \hline
		230920/1                                       & 1.6                     & 6.0                     & 0.7                     & -                       & -    \\
		230921/2                                       & 1.4                     & 6.4                     & 0.9                     & -                       & -    \\
		231007/4                                       & 9.5                     & 12.1                    & 4.7                     & -                       & -    \\
		231105/6                                       & 6.1                     & 1.6                     & 1.6                     & 8.9                     & 0.7  \\
		231105\_aft/2                                  & 1.8                     & 1.5                     & 3.2                     & 0.8                     & 0.6  \\ \hline
		230920/2                                       & 156.9                   & 188.3                   & 58.9                    & -                       & -    \\
		230921/3                                       & 200.3                   & 202.6                   & 134.2                   & -                       & -    \\
		230921/5                                       & 174.6                   & 194.7                   & 58.7                    & -                       & -    \\
		231007/2                                       & 200.5                   & 194.2                   & 68.4                    & -                       & -    \\
		231019/1                                       & 64.7                    & 197.4                   & 58.2                    & -                       & -    \\
		231105/2                                       & 37.8                    & 143.9                   & 31.7                    & 137.0                   & 25.8 \\
		231105/3                                       & 118.3                   & 134.5                   & 62.0                    & 102.8                   & 33.5 \\
		231105\_aft/4                                  & 167.8                   & 125.5                   & 72.4                    & 69.0                    & 9.1  \\
		231109/3                                       & 190.2                   & 128.3                   & 209.2                   & 61.0                    & 11.6 \\ \hline
		230921/4                                       & 27.5                    & 58.3                    & 12.6                    & -                       & -    \\
		231019/2                                       & 35.4                    & 57.6                    & 59.1                    & -                       & -    \\
		231105/4                                       & 18.2                    & 25.8                    & 2.4                     & 37.5                    & 17.9 \\
		231105/5                                       & 26.6                    & 27.9                    & 52.6                    & 39.4                    & 15.1 \\
		231105\_aft/5                                  & 32.1                    & 16.8                    & 51.2                    & 20.9                    & 10.6 \\
		231109/4                                       & 37.3                    & 31.2                    & 45.7                    & 22.0                    & 16.3 \\ \hline
		231208/4                                       & 75.8                    & 60.7                    & 8.5                     & 43.4                    & 5.7  \\
		231213/4                                       & 4.9                     & 80.5                    & 30.8                    & 47.4                    & 2.0  \\
		231213/5                                       & 11.6                    & 64.7                    & 22.7                    & 48.2                    & 2.8  \\
		240115/3                                       & 25.9                    & 46.9                    & 3.1                     & 52.5                    & 5.0  \\
		240116/5                                       & 42.4                    & 75.7                    & 32.7                    & 56.4                    & 4.4  \\
		240116\_eve/5                                  & 18.5                    & 57.9                    & 23.3                    & 51.7                    & 3.2  \\
		240123/3                                       & 58.4                    & 53.6                    & 22.3                    & 53.3                    & 4.2  \\ \hline
		231201/2                                       & 166.5                   & 117.4                   & 41.4                    & 85.0                    & 1.2  \\
		231201/3                                       & 92.9                    & 53.9                    & 44.5                    & 94.7                    & 9.3  \\
		231208/5                                       & 129.7                   & 116.5                   & 54.5                    & 28.8                    & 1.0  \\
		231213/2                                       & 134.5                   & 187.9                   & 2.1                     & 180.1                   & 7.4  \\
		231213/3                                       & 129.6                   & 210.8                   & 100.5                   & 119.2                   & 23.2 \\
		240113/2                                       & 184.3                   & 182.1                   & 31.1                    & 175.2                   & 77.1 \\
		240113/3                                       & 101.8                   & 105.2                   & 77.7                    & 63.8                    & 1.3  \\
		240116\_eve/4                                  & 145.9                   & 186.5                   & 96.0                    & 179.5                   & 11.9 \\ \hline
		240113/5                                       & 171.2                   & 138.3                   & 116.8                   & 58.6                    & 1.5  \\
		240115/2                                       & 159.7                   & 208.8                   & 79.7                    & 114.7                   & 1.0  \\
		240116/4                                       & 169.4                   & 170.5                   & 28.0                    & 216.7                   & 1.3  \\ \hline
		231208/1                                       & 2.1                     & 8.1                     & 4.7                     & 5.1                     & 0.6  \\
		231213/1                                       & 1.7                     & 7.5                     & 1.7                     & 5.9                     & 0.8  \\
		240113/1                                       & 32.6                    & 15.8                    & 3.2                     & 12.4                    & 0.5  \\ \hline
		240116/2                                       & 69.6                    & 76.6                    & 68.4                    & 78.5                    & 0.2  \\
		240116\_eve/3                                  & 76.5                    & 76.6                    & 63.7                    & 76.9                    & 0.2  \\
		240123/2                                       & 8.6                     & 73.6                    & 31.3                    & 73.1                    & 0.3  \\ \hline
	\end{tabular}
\end{table}

\subsection{Radar place recognition}
Likewise, numerous radar-based place recognition methods have been developed over the years.
For spinning radars, notable approaches include k-Radar++ \citep{demartiniKRadarCoarsetoFineFMCW2020}, Kidnapped Radar \citep{saftescuKidnappedRadarTopological2020}, Open-RadVLAD \citep{gaddOpenRadVLADFastRobust2024}, ReFeree \citep{kimReFereeRadarBasedLightweight2024}, and RaPlace \citep{jangRaPlacePlaceRecognition2023}.
These methods primarily work with range-azimuth data.

In contrast, 4D single-chip radars provide additional Doppler information and vertical data, offering the potential for improved place recognition.
Several learning-based methods utilizing single-chip radars have been proposed,
including AutoPlace \citep{caiAutoPlaceRobustPlace2022} designed for 3D radars, voxel-based TransLoc4D \citep{pengTransLoc4DTransformerbased4D2024},
and SPR for single sparse scans \citep{herraezSPRSingleScanRadar2024},
and 4D RadarPR that adaptively fuses point features and multiscale context features \citep{chen4DRadarPRContextaware2025}.

In this work, we evaluate three radar-based place recognition methods using either ARS548 or Eagle radar data from our dataset.
These methods include the classic point-based PointNetVLAD \citep{uyPointNetVLADDeepPoint2018}, 
the projection-based AutoPlace \citep{caiAutoPlaceRobustPlace2022},
and TransLoc4D \citep{pengTransLoc4DTransformerbased4D2024},
all of which were retrained from scratch using each type of radar data in our dataset.

To begin, the dataset is divided into three subsets: training, validation, and test.
The training set consists of seven sequences from four routes: Starlake, Software School, Info Faculty, and August 1 Road, captured under various environmental conditions.
The validation set, used to assess the model's effectiveness during training, includes four sequences from both the Info and Arts Faculty and the Info Faculty routes.
The remaining sequences, which are not part of the training or validation sets, form the test set.
These test set includes some sequences from routes unseen in the training data.
In both the validation and test sets, there are multiple sequences for a route.
For these, the first sequence serves as the database, while the others act as queries.
During training, thresholds for positive and negative samples were set to 9 m and 18 m, respectively,
with a 9 m threshold applied for positive samples during testing.
Additionally, given the radar’s approximate 120$^\circ$ field of view, the heading angle difference threshold was
set to 75$^\circ$ for training and 30$^\circ$ for testing to keep positive candidates.

For data preparation, moving objects were filtered using the GNC method as described in \cite{zhuang4DIRIOM4D2023}.
Frames were then selected at intervals of five for place recognition.
For the Eagle radar, enhanced accumulative point cloud frames were used.
To address the sparsity of the ARS548 point clouds, 
a window of seven frames centered on the selected frames was used to form an aggregated frame, 
with compensation for relative motion derived from the reference trajectory.

For all three methods, we primarily use the default settings from the public codebases unless otherwise specified.
For PointNetVlad, the input consists of 4D tuples of point coordinates and RCS, and the output is a 256D descriptor.
The model was trained for 40 epochs using the lazy quadruplet loss with a learning rate of $1\cdot 10^{-5}$ and a batch size of 2.
Each batch includes 1 query frame, 2 positive frames, 18 negative frames, and 1 hard negative frame.
For AutoPlace, the model was trained for 200 epochs with an initial learning rate of $1\cdot 10^{-4}$ and a batch size of 16.
For TransLoc4D, the input consists of 5D tuples ($\mathbf{p}$, $s$, $I$), 
where $\mathbf{p}$ is radar point coordinates, 
$s$ is the relative azimuth angle,
and $I$ is the RCS, and the output is also 256D.
The model was trained for 500 epochs with an initial learning rate of $1\cdot 10^{-3}$ and a batch size of 768.

The recall rates for the three methods are listed in Table~\ref{tab:radar-place}.
Despite its age, PointNetVlad, originally developed for lidar, often produced the best place recognition results.
The relatively low recall in the top five candidates using the Eagle radar may be attributed to motion distortion in the accumulative radar point clouds.
The suboptimal performance of the recent TransLoc4D method may be attributed to its reliance on velocity directions for generating a place descriptor, 
which vary across different traversals \citep{herraezSPRSingleScanRadar2024}.

\begin{table*}[]
	\centering
	\caption{Recall of three recent place recognition methods on the proposed dataset, using the Continental ARS548 and Oculii Eagle enhanced 4D radar data.
		The three methods include the classic point-based PointNetVlad, the projection-based AutoPlace, and the voxel-based TransLoc4D.
		r@K represents the recall when considering the top K retrieved samples.
		K can also be set as 1\% of the total number of descriptors in the database.
		The route abbreviations follow Table~\ref{tab:seqs}.
	}
	\label{tab:radar-place}
	\begin{tabular}{cc|cccc|cccc}
		\hline
		\multirow{2}{*}{Routes}        & \multirow{2}{*}{Methods} & \multicolumn{4}{c|}{Continental ARS548}                                                                   & \multicolumn{4}{c}{Oculii Eagle enhanced}                                                                \\ \cline{3-10} 
		&                          & \multicolumn{1}{l}{r@1} & \multicolumn{1}{l}{r@5} & \multicolumn{1}{l}{r@10} & \multicolumn{1}{l|}{r@1\%} & \multicolumn{1}{l}{r@1} & \multicolumn{1}{l}{r@5} & \multicolumn{1}{l}{r@10} & \multicolumn{1}{l}{r@1\%} \\ \hline
		\multirow{3}{*}{\textbf{bc}}   & PointNetVlad             & 81.0                    & 93.6                    & 98.9                     & 100.0                      & 72.2                    & 82.7                    & 87.6                     & 76.5                      \\
		& AutoPlace                & 76.8                    & 90.2                    & 93.9                     & 88.6                       & 71.3                    & 83.5                    & 89.6                     & 75.4                      \\
		& TransLoc4D               & 35.9                    & 74.6                    & 84.7                     & 91.5                       & 49.4                    & 69.5                    & 76.1                     & 84.1                      \\ \hline
		\multirow{3}{*}{\textbf{sl}}   & PointNetVlad             & 59.1                    & 79.0                    & 85.7                     & 89.7                       & 64.8                    & 76.5                    & 81.0                     & 86.6                      \\
		& AutoPlace                & 43.6                    & 60.8                    & 68.0                     & 72.2                       & 47.6                    & 58.6                    & 64.9                     & 80.2                      \\
		& TransLoc4D               & 21.9                    & 45.8                    & 56.9                     & 70.1                       & 49.2                    & 68.5                    & 75.4                     & 89.5                      \\ \hline
		\multirow{3}{*}{\textbf{ss}}   & PointNetVlad             & 98.3                    & 99.8                    & 100.0                    & 100.0                      & 57.9                    & 70.1                    & 76.3                     & 80.2                      \\
		& AutoPlace                & 97.5                    & 99.7                    & 100.0                    & 100.0                      & 41.2                    & 52.4                    & 59.7                     & 65.7                      \\
		& TransLoc4D               & 88.8                    & 96.9                    & 97.7                     & 99.0                       & 46.1                    & 65.0                    & 73.6                     & 83.4                      \\ \hline
		\multirow{3}{*}{\textbf{if}}   & PointNetVlad             & 97.8                    & 99.4                    & 99.6                     & 99.7                       & 95.9                    & 97.7                    & 98.0                     & 98.2                      \\
		& AutoPlace                & 95.4                    & 98.3                    & 99.0                     & 99.4                       & 88.7                    & 94.9                    & 96.4                     & 96.6                      \\
		& TransLoc4D               & 90.7                    & 98.6                    & 99.3                     & 99.6                       & 86.5                    & 96.7                    & 98.1                     & 98.8                      \\ \hline
		\multirow{3}{*}{\textbf{iaf}}  & PointNetVlad             & 96.9                    & 99.0                    & 99.3                     & 99.6                       & 92.4                    & 96.6                    & 97.7                     & 98.8                      \\
		& AutoPlace                & 96.2                    & 98.6                    & 99.2                     & 99.3                       & 85.3                    & 92.4                    & 95.1                     & 97.7                      \\
		& TransLoc4D               & 91.0                    & 98.1                    & 99.0                     & 99.6                       & 80.5                    & 94.5                    & 97.2                     & 98.8                      \\ \hline
		\multirow{3}{*}{\textbf{iaef}} & PointNetVlad             & 97.3                    & 99.1                    & 99.4                     & 99.7                       & 87.9                    & 93.5                    & 95.4                     & 97.3                      \\
		& AutoPlace                & 95.6                    & 97.6                    & 98.2                     & 99.2                       & 81.8                    & 89.1                    & 92.7                     & 96.7                      \\
		& TransLoc4D               & 92.6                    & 98.0                    & 98.6                     & 99.5                       & 73.0                    & 87.3                    & 90.2                     & 94.1                      \\ \hline
		\multirow{3}{*}{\textbf{st}}   & PointNetVlad             & 100.0                   & 100.0                   & 100.0                    & 100.0                      & 98.5                    & 99.6                    & 100.0                    & 100.0                     \\
		& AutoPlace                & 99.7                    & 99.8                    & 99.9                     & 100.0                      & 93.3                    & 96.4                    & 98.0                     & 96.1                      \\
		& TransLoc4D               & 99.4                    & 100.0                   & 100.0                    & 100.0                      & 91.7                    & 96.3                    & 97.1                     & 98.7                      \\ \hline
		\multirow{3}{*}{\textbf{81r}}  & PointNetVlad             & 93.1                    & 97.0                    & 97.8                     & 98.8                       & 87.7                    & 94.7                    & 96.5                     & 97.6                      \\
		& AutoPlace                & 94.3                    & 98.1                    & 99.0                     & 99.8                       & 85.2                    & 92.3                    & 94.4                     & 96.8                      \\
		& TransLoc4D               & 89.8                    & 97.5                    & 98.7                     & 99.9                       & 73.4                    & 91.0                    & 95.3                     & 98.7                      \\ \hline
	\end{tabular}
\end{table*}

\section{Conclusions}
\label{sec:conclusions}
This paper introduces a large-scale 4D radar dataset designed for localization and mapping applications leveraging multi-sensor fusion.
The dataset was collected using three platforms across diverse environmental conditions, including rainy days, nighttime, campus roads, and tunnels.
In total, 44 sequences were repeatedly recorded over eight distinct routes.

Accurate reference poses (at the start and end of large-scale sequences) were generated using a LIO method
that sequentially localizes undistorted Hesai lidar frames to a TLS map.
A data inversion technique enables backward LIO processing, 
which in turn allows two-way reference trajectory smoothing and quality assessment of the resulting reference trajectories.

Sensor data synchronization was achieved using a two-step scheme.
First, hardware-synchronized lidar data was used to map message host times to GNSS times, eliminating jitter and long-term drift.
Second, odometry and correlation algorithms were employed to estimate constant time offsets between the sensor data and the GNSS/INS solution.

Using this dataset, we evaluated several recent radar-based odometry and place recognition methods,
showing the challenges inherent to 4D radar data.
Additionally, as the dataset provides dense reference point clouds, it is well-suited for evaluating neural 3D reconstruction methods \citep{mingBenchmarking2025}.

\begin{acks}
We acknowledge the assistance of Xiaochen Wang and Hanwen Qi in collecting the TLS data with the RTC360, lent by Prof. Xinlian Liang.
We are also grateful to Shangjia Liu and Qi Liu for their help with data collection,
to Dr. Yusheng Wang, Jian Zhou, and Hailiang Tang for fruitful discussions, to Prof. Charles Toth and the anonymous reviewers for their helpful and stimulating suggestions. 
\end{acks}

\begin{funding}
This research is partly funded by the National Key Research and Development Program of China (International Scientific and Technological Cooperation Program) under Grant 2022YFE0139300,
the National Natural Science Foundation of China under Grant 42374047, and 
the Fundamental Research Fund Program of LIESMARS.
\end{funding}

\theendnotes

\bibliographystyle{SageH}
\bibliography{zotero.bib}

\end{document}